\documentclass[sigconf]{acmart}
\usepackage{colortbl}
\usepackage{multirow}
\usepackage{subfig}
\usepackage{fancyhdr}

\AtBeginDocument{%
  }
\renewcommand\footnotetextcopyrightpermission[1]{}
\acmConference[MM'25]{the 33rd ACM International Conference on Multimedia}
{October 27--31, 2025}{Dublin, Ireland}

\begin{document}

\title{Butter: Frequency Consistency and Hierarchical Fusion for Autonomous Driving Object Detection}




\author{Xiaojian Lin}
\email{chrislim@connect.hku.hk}
\affiliation{%
\institution{Tsinghua University}
\city{Beijing}
\country{China}}

\author{Wenxin Zhang}
\email{zwxzhang12@163.com}
\affiliation{
\institution{University of Chinese Academy of Sciences}
\city{Beijing}
\country{China}}

\author{Yuchu Jiang}
\email{kamichanw@seu.edu.cn}
\affiliation{%
\institution{Southeast University}
\city{Nanjing}
\country{China}}

\author{Wangyu Wu}
\email{Wangyu.wu@liverpool.ac.uk}
\affiliation{%
\institution{University of Liverpool}
\city{Liverpool}
\country{UK}}

\author{Yiran Guo}
\email{only1gyr@gmail.com}
\affiliation{%
\institution{Beijing Institute of Technology}
\city{Beijing}
\country{China}}

\author{Kangxu Wang}
\email{kx-wang23@mails.tsinghua.edu.cn}
\affiliation{%
\institution{Tsinghua University}
\city{Beijing}
\country{China}}

\author{Zongzheng Zhang}
\email{zzongzheng0918@gmail.com}
\affiliation{%
\institution{Tsinghua University}
\city{Beijing}
\country{China}}

\author{Guijin Wang}
\email{wangguijin@tsinghua.edu.cn}
\affiliation{%
\institution{Tsinghua University}
\city{Beijing}
\country{China}}

\author{Lei Jin$^\dag$}
\email{jinlei@bupt.edu.cn}
\affiliation{%
\institution{Beijing University of Posts and Telecommunications}
\city{Beijing}
\country{China}}

\author{Hao Zhao$^\dag$}
\email{zhaohao@air.tsinghua.edu.cn}
\affiliation{%
\institution{Tsinghua University}
\city{Beijing}
\country{China}}







\begin{abstract}
  
Hierarchical feature representations play a pivotal role in computer vision, particularly in object detection for autonomous driving. Multi-level semantic understanding is crucial for accurately identifying pedestrians, vehicles, and traffic signs in dynamic environments. However, existing architectures, such as YOLO and DETR, struggle to maintain feature consistency across different scales while balancing detection precision and computational efficiency. To address these challenges, we propose Butter, a novel object detection framework designed to enhance hierarchical feature representations for improving detection robustness. Specifically, Butter introduces two key innovations: Frequency-Adaptive Feature Consistency Enhancement (FAFCE) Component, which refines multi-scale feature consistency by leveraging adaptive frequency filtering to enhance structural and boundary precision, and Progressive Hierarchical Feature Fusion Network (PHFFNet) Module, which progressively integrates multi-level features to mitigate semantic gaps and strengthen hierarchical feature learning. Through extensive experiments on BDD100K, KITTI, and Cityscapes, Butter demonstrates superior feature representation capabilities, leading to notable improvements in detection accuracy while reducing model complexity. By focusing on hierarchical feature refinement and integration, Butter provides an advanced approach to object detection that achieves a balance between accuracy, deployability, and computational efficiency in real-time autonomous driving scenarios. Our model and implementation are publicly available at
\href{https://github.com/Aveiro-Lin/Butter}{https://github.com/Aveiro-Lin/Butter}, facilitating further research and validation within the autonomous driving community.

\end{abstract}





\maketitle


\pagestyle{fancy}
\fancyhf{}
\fancyhead[C]{MM'25, October 27–31, 2025, Dublin, Ireland}

\makeatletter
\let\ps@plain\ps@fancy
\makeatother

\renewcommand{\thefootnote}{}
\footnote{\textsuperscript{\dag} are equally corresponding authors. \\ The paper is supported by National Natural Science Foundation of China No.62472046.}

\begin{figure}[t]
\begin{center}
   \includegraphics[width=1\columnwidth]{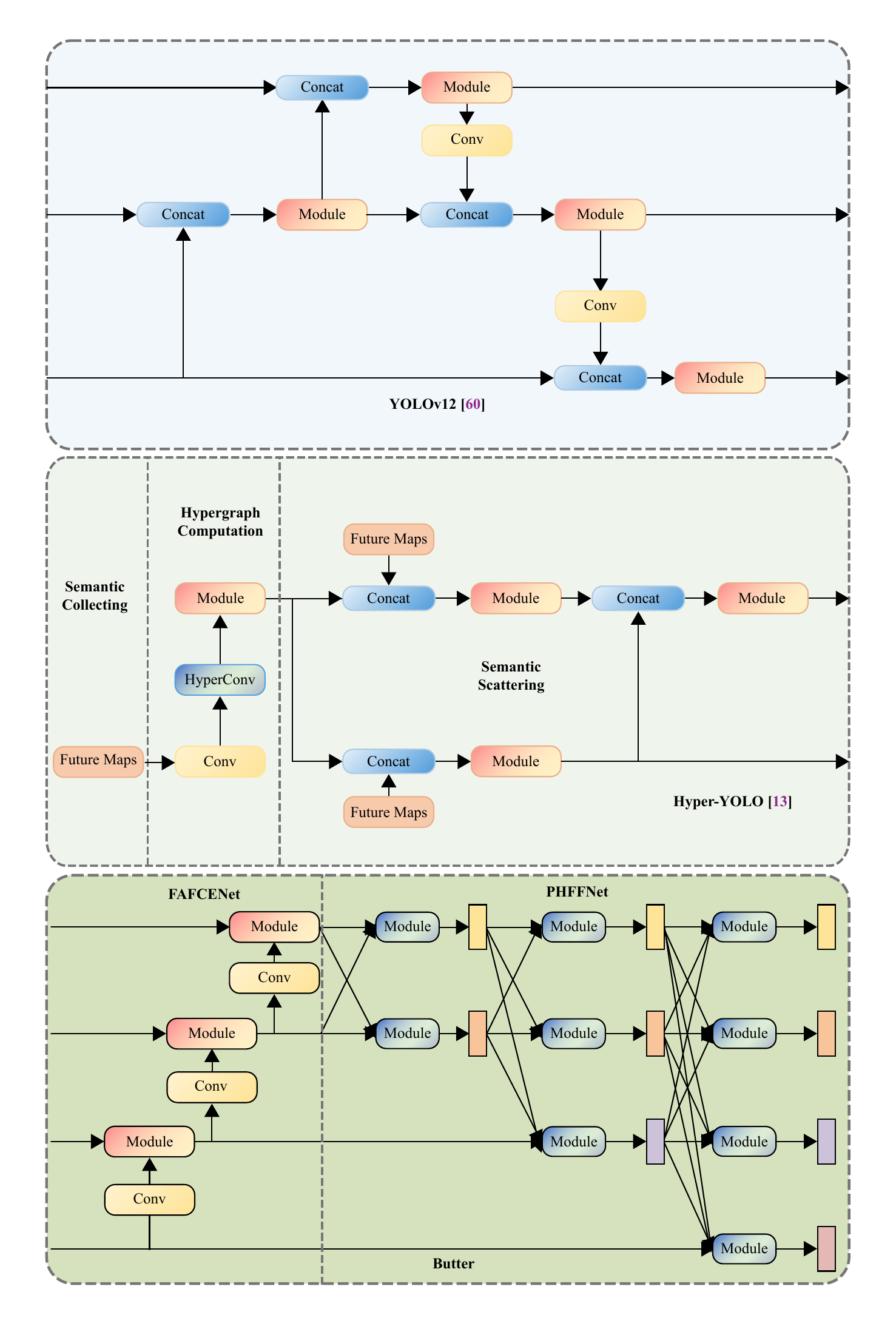}
\end{center}
\vspace{-10pt}
\caption{Comparison between the neck of proposed Butter and other 2 previous most popular 2D object detection methods YOLOv12~\cite{Tian2025} and Hyper-YOLO~\cite{hyperyolo}. }
\label{fig:Compare with Previous}
\end{figure}

\section{Introduction}

Recent advances in autonomous driving technology have made object detection essential for intelligent transportation systems. This technology allows for real-time detection of objects such as pedestrians, cyclists, and traffic signs, which is crucial for decision-making~\cite{mahaur2023small, wang2025physically, benjumea2021yolo, wang2025unified, zhang2025self}. Despite these advancements, challenges exist due to the complexity of traffic scenarios and diverse targets. The need for real-time performance exposes limitations in the accuracy, robustness, and real-time capabilities of current algorithms in dynamically changing environments~\cite{chen2021deep, wang2023banet}.

Object detection algorithms in deep learning are typically categorized into two types: two-stage and one-stage algorithms. One-stage detectors, such as YOLO (You Only Look Once) models~\cite{glenn2023yolov8,Wang2024v10,Khanam2024v11,Tian2025,hyperyolo,wang2023gold,wang2024yolov9} and SSD~\cite{liu2016ssd}, are known for their high-speed detection capabilities. In contrast, two-stage detectors like R-CNN~\cite{girshick2016r-cnn} and faster R-CNN~\cite{he2017faster} tend to offer higher detection accuracy. Although two-stage models excel in accuracy, one-stage models are preferred for their faster processing times. In response to the real-time demands of autonomous driving scenarios, we focus lightweight approaches, which fall into the category of one-stage object detection.

Current novel object detection algorithms, such as Hyper-YOLO~\cite{hyperyolo} and YOLOv12~\cite{Tian2025}, have demonstrated strong performance in detecting large objects in autonomous driving. However, they still face challenges when it comes to handling complex backgrounds, occlusions, and objects with varying scales~\cite{zamanakos2021comprehensive,ding2024hint}. This is especially evident in traffic environments, where factors like differences in object sizes, changes in lighting, and background noise further complicate detection tasks~\cite{gupta2021deep,huang2025uniformity,zhang2025detect}. Moreover, effectively detecting objects in autonomous driving systems, which require high real-time performance, remains a significant challenge for existing technologies~\cite{bachute2021autonomous,tilijani2023optimized}. To address this, object detection algorithms must not only excel at precise feature extraction, but also be optimized for computational efficiency and designed to be lightweight, ensuring that they meet the real-time and high-efficiency demands of autonomous driving systems while being easy to deploy~\cite{liu2021ground,subramanian2023mdho}.

To achieve a balance between lightweight design and detection accuracy, we propose a novel 2D real-time object detection model, Butter. This model achieves real-time object detection with high accuracy while maintaining a low parameter count for efficient deployment. It reduces computational overhead, making it ideal for autonomous driving scenarios. 
As shown in Fig.~\ref{fig:Compare with Previous}, unlike the latest advanced models such as Hyper-YOLO~\cite{hyperyolo} and YOLOv12~\cite{Tian2025}, our model incorporates two modules in the neck: the Frequency-Adaptive Feature Consistency Enhancement (FAFCE) and the Context-Aware Spatial Fusion (CASF). These modules optimize feature alignment and reduce semantic gaps between feature levels, thereby enhancing category consistency and boundary precision while improving the multi-scale feature representation capability of the object detection model. 
In parallel, recent works have explored using language priors to enhance feature expressiveness and open-set robustness in road anomaly detection~\cite{tian2023unsupervised}.
In experiments, our lightweight Butter method with low parameters achieves accurate detection on 3 autonomous driving datasets. 

Our main contributions are summarized as follows.
\begin{itemize}

   \item We propose a FAFCENet component, which employs contextual low-frequency damping to refine semantic representations, contextual high-frequency amplifier to restore object boundaries, and a feature resampling module to improve spatial consistency, thereby enhancing category coherence and boundary precision for robust object localization and classification.
   
   \item We propose PHFFNet, which progressively integrates hierarchical feature aggregation to enhance semantic representations. It employs CASF to optimize feature alignment, reducing semantic gaps across hierarchical levels and enhancing multi-scale feature representation in object detection.

   \item Butter addresses the limitations of existing object detection methods, particularly in integrating multi-scale features within the neck layer. The model achieves state-of-the-art (SOTA) results with fewer than 10 million parameters on datasets like KITTI~\cite{geiger2012ready}, Cityscapes~\cite{cityscapes}, and BDD100K~\cite{yu2020bdd100k}. For instance, on the Cityscapes~\cite{cityscapes} dataset, Butter reduces the parameter count by over 40\% while improving mAP@0.5 by 1.8\%, showcasing significant advances in both detection accuracy and parameter-efficiency.

\end{itemize}

\section{Related Works}

\subsection{Frequency-Based Feature Refinement.}

Frequency analysis, a fundamental technique in traditional signal processing~\cite{pitas2000digital}, has proven valuable in deep learning, especially for computer vision. It has been used to investigate generalization abilities~\cite{wang2020high} and optimization techniques~\cite{yin2019fourier} in Deep Neural Networks (DNNs). Manipulating high-frequency components enables adversarial attacks~\cite{luo2022frequency,yin2019fourier} and weaken feature representation by reducing intra-category similarity~\cite{luo2022frequency}. Studies by Rahaman et al.~\cite{rahaman2019spectral} and Xu et al.~\cite{xu2021deep} show that DNNs prioritize low-frequency patterns during training, a phenomenon known as spectral bias. Zhang et al.~\cite{zhang2019making} examine the effects of frequency aliasing on translation invariance, while FLC~\cite{grabinski2022frequencylowcut} confirms that aliasing reduces model robustness. Low-pass filters, as demonstrated by Chen et al.~\cite{chen2023instance}, can suppress high-frequency noise, particularly in low-light conditions, and AdaBlur~\cite{zou2023delving} mitigates aliasing by applying content-aware low-pass filtering during downsampling. Further optimization, leveraging additional frequency components, has shown to improve performance~\cite{qin2021fcanet,magid2021dynamic}, particularly through discrete cosine transform coefficients for channel attention mechanisms. Traditional convolutional theorems in DNNs~\cite{huang2023adaptive} use adaptive frequency filters to act as global label mixers, and various frequency-domain methods have been integrated into DNNs to enhance non-local feature learning~\cite{huang2023adaptive,rao2021global,li2021fourier}.

Current frequency analysis improves neural network generalization, adversarial robustness, and feature extraction but focuses on single-frequency operations, neglecting frequency consistency in multiscale fusion. In contrast, the FAFCE component integrates low-frequency damping, high-frequency amplifier, and displacement calculator to ensure frequency consistency across levels, adapting to data variations, reducing information loss, and enhancing multiscale detection and localization.

\subsection{Feature Fusion.}

Feature fusion enhances both semantic and spatial representations by integrating low- and high-resolution features. This is achieved through top-down approaches such as DeepLabv3+~\cite{chen2018encoder} and U-Net~\cite{falk2019unet}, as well as bottom-up approaches like SeENe~\cite{pang2019towards} and DLA~\cite{yu2018deep}. However, simple fusion methods struggle with resolution and semantic inconsistencies. Modern techniques address these challenges through sampling-based methods, such as SFNet~\cite{li2023sfnet}, FaPN~\cite{huang2021fapn}, and AlignSeg~\cite{huang2021alignseg}, which align features using spatial offsets. They also include kernel-based methods, such as deconvolution~\cite{zeiler2014visualizing}, Pixel Shuffle~\cite{shi2016real}, A2U~\cite{dai2021learning}, and CARAFE~\cite{wang2021carafe}, which upscale features using fixed or learnable kernels.
Fusion modules like CARAFE~\cite{wang2021carafe}, ASFF~\cite{liu2019learning}, and DRFPN~\cite{ma2020dual} improve multi-scale feature fusion within FPN, enhancing detection accuracy~\cite{jin2024tod3cap}.

FPN-based methods~\cite{lin2017feature} improve object detection by predicting multi-level features, but they struggle to integrate high- and low-level features effectively. To address this, PANet~\cite{liu2018path} adds a bottom-up pathway, and NASFPN~\cite{ghiasi2019nasfpn} uses neural architecture search to optimize feature connections. Zhao et al.~\cite{zhao2019new}, Wu et al.~\cite{wu2023rornet}, and Ma et al.~\cite{ma2023explore} propose alternative optimization methods. While GraphFPN~\cite{zhao2021graphfpn} and FPT~\cite{zhang2020fpt} employ graph neural networks and self-attention~\cite{li2022toist} for feature exchange and aggregation, respectively, they increase computational complexity. In contrast, PHFFNet uses standard convolutions for a more efficient solution, making it more suitable for real-world applications. Similar to multimodal transformers in embodied AI~\cite{li2023understanding}, our design emphasizes structured feature alignment with minimal overhead.

\begin{figure*}[t]
   \centering
   \includegraphics[width=0.9\textwidth]{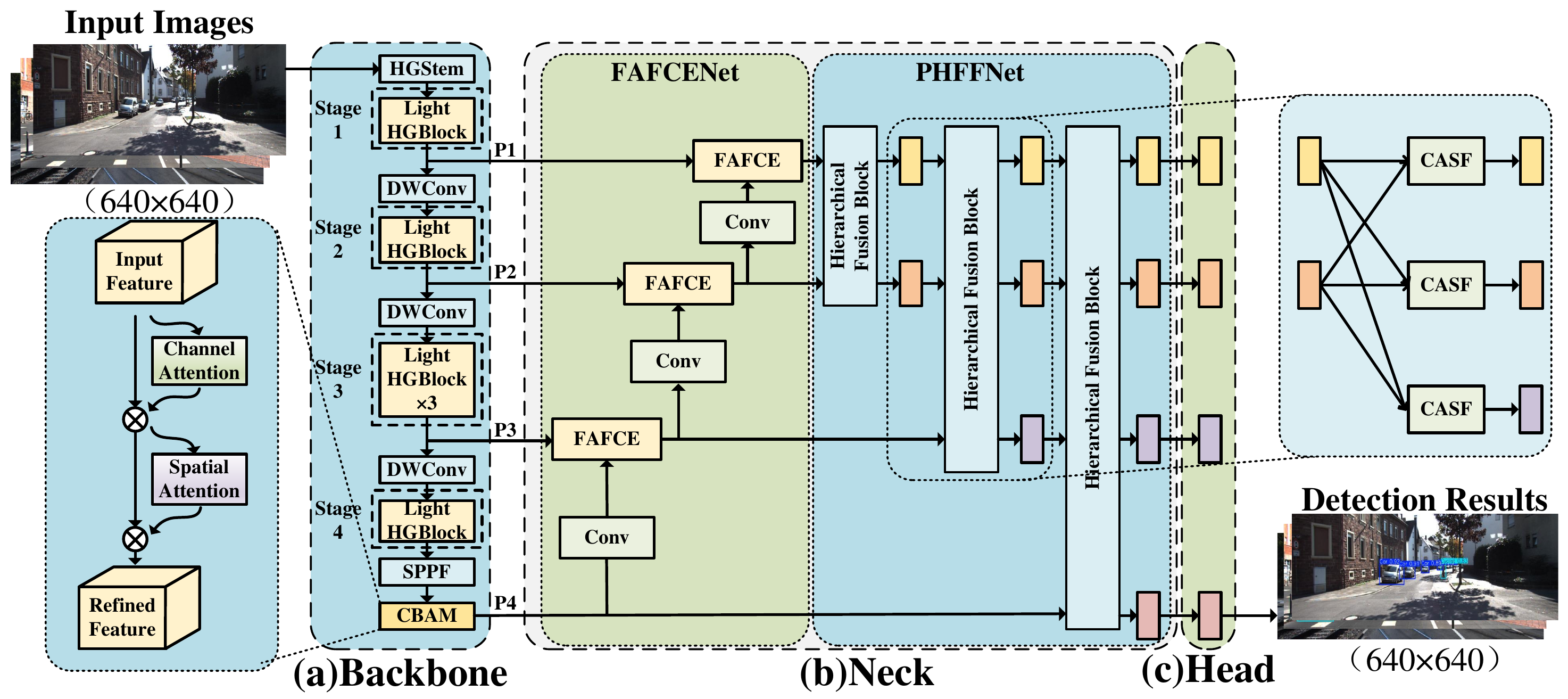}
   \vspace{-10pt}
\setlength{\abovecaptionskip}{0.5cm}
   \caption{Comprehensive workflow of the Butter model for autonomous driving object detection.
(1) The workflow begins with a monocular image of $640\times640$ pixels processed through the HGStem of the Backbone to extract features. These features are subsequently refined through a series of lightweight HGBlocks, Depthwise Convolutions (DWConv)~\cite{Howard2017MobileNets}, and Convolutional Block Attention Modules (CBAM)~\cite{Woo2018CBAM} before entering the Neck module. The Neck consists of two parts: FAFCENet and PHFFNet. Following the Neck, the model utilizes four heads in the Head layer to produce outputs, including the original image annotated with class labels, confidence scores, and bounding boxes. 
(2) The CBAM module in the lower left corner applies channel and spatial attention to guide the network toward informative features.
(3) The Hierarchical Fusion Block in the upper right corner enables multi-level feature interaction within the Context-Aware Spatial Fusion (CASF) module. Horizontal arrows represent feature exchange, while only diagonal arrows are used to indicate upsampling and downsampling.
}
\label{fig:overview of architecture}
\end{figure*}

Current feature fusion methods improve high-low feature integration but struggle with cross-level inconsistency, boundary loss, and high computational cost. PHFFNet uses hierarchical aggregation and CASF for efficient integration. The FAFCE component integrates adaptive damping, high-frequency amplification, and feature resampling to improve spatial and frequency consistency, thereby enhancing feature representation and boosting multi-scale detection performance.

\section{Method}
\subsection{Overview}

\noindent\textbf{Task Description.} 
Object detection in autonomous driving is crucial for identifying various objects, such as pedestrians, vehicles, and traffic signs, from monocular RGB images in complex, dynamic environments. We present Butter, a novel object detection method specifically designed for autonomous driving scenarios.


\noindent\textbf{Overall Framework.} As shown in Fig.\ref{fig:overview of architecture}, Butter consists of three primary branches. The \textbf{Backbone Branch} (Fig.\ref{fig:overview of architecture} (a)) leverages a lightweight version of the HGNetV2 architecture, enhanced with lightweight Depthwise Convolutions (DWConv)~\cite{Howard2017MobileNets} and Convolutional Block Attention Modules (CBAM)~\cite{Woo2018CBAM} Module. This branch processes input monocular $640\times640$ RGB images to extract essential features, improving both computational efficiency and detection accuracy.  The \textbf{Neck Branch} (Fig.~\ref{fig:overview of architecture} (b)) consists of two crucial elements: \textbf{FAFCE} as a component and \textbf{PHFFNet} as a module. The \textbf{FAFCE} component enhances the model’s ability to maintain feature consistency and boundary precision. At the same time, the \textbf{PHFFNet} module integrates low-level and high-level features progressively, helping to bridge semantic gaps and improve the accuracy of object detection. The \textbf{Head Branch} (Fig.~\ref{fig:overview of architecture} (c)) distinguishes itself from traditional YOLOv12~\cite{Tian2025} models by using four detection heads instead of the typical three. 

\label{subsection:overview}
\subsection{Butter Method}

\noindent\textbf{Backbone Branch.} HGNetV2 is short for PP-HGNetV2, a new network model developed by Baidu PaddlePaddle Vision Team. Due to its excellent real-time performance and accuracy, it has shown outstanding results in tasks such as single- or multilabel classification, object detection, and semantic segmentation. We use it as the baseline for our backbone model.

\begin{figure}[t]
\begin{center}
   \includegraphics[width=1\columnwidth]{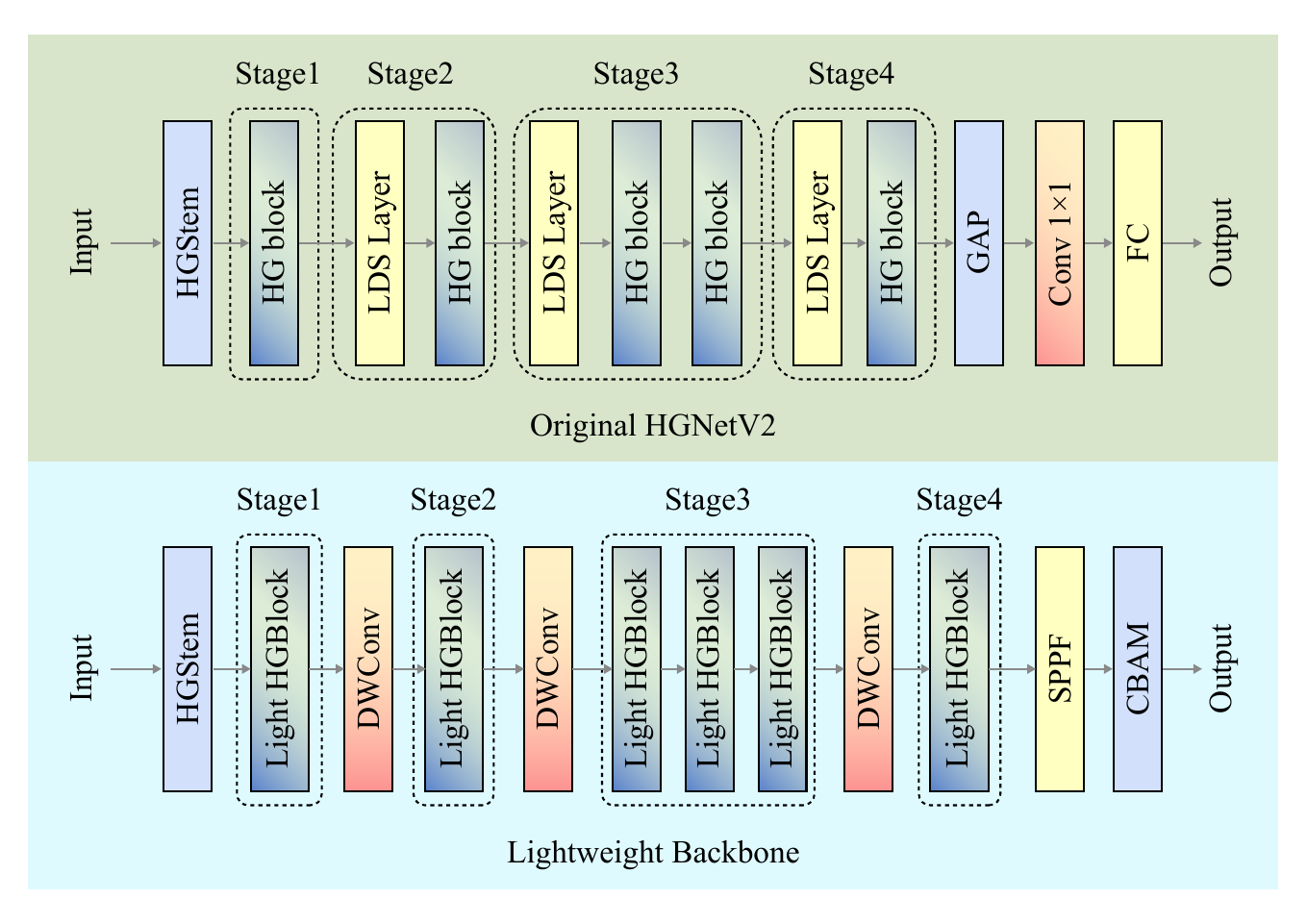}
\end{center}
\vspace{-10pt}
\caption{Architecture comparison between the original HGNetV2 and the lightweight backbone of Butter.}
\label{fig:backbone compare}
\end{figure}

We make lightweight improvements based on the original HGNetV2, as illustrated in Fig.~\ref{fig:backbone compare}, the first step of Butter involves processing monocular RGB images through an HGStem to extract features. These features are then passed through the first stage of the lightweight HGBlock. The main difference between the lightweight HGBlock we propose and the traditional HGBlock is that we replace the convolutional layers with lightweight layers such as GhostConv, RepConv, DWConv, and LightConv, thereby reducing the parameter count of the model's backbone. Compared to the traditional HGNetV2, we have made the following optimizations. We replace the LDS (Learnable Down-sampling) operation that was originally applied starting from Stage 2-4 with DWConv. The DWConv has been widely applied in various object detection models. Additionally, after Stage 4, we replace the Global Average Pooling (GAP) with Spatial Pyramid Pooling Fast (SPPF) and substitute the Fully Connected (FC) layer with CBAM. These adjustments improve the model's computational efficiency and enhance its feature extraction and attention mechanisms. 
The introduction of SPPF and CBAM into the backbone, rather than the neck or head, leverages the role of the backbone in the early extraction of features. In autonomous driving, optimizing feature quality and multi-scale information at this stage is critical for perceiving complex environments. By placing these modules here, we enhance the features' discriminative power, improving downstream tasks like object localization and classification. This strategy boosts overall network performance, ensuring robust feature extraction for accurate and efficient real-time detection in driving scenarios.

The decision to add SPPF and CBAM after Stage 4, rather than earlier, stems from the distinct roles of each stage in object detection. Stages 1 to 3 focus on extracting basic and fine-grained features, emphasizing low-level structures from the input image, without considering object scales or specific feature selection. In autonomous driving, this early extraction is critical for fast, accurate detection of objects in dynamic environments. Introducing SPPF and CBAM at this stage would add unnecessary computational load and disrupt the learning of low-level features. Stage 4, being the final stage, contains rich contextual information and multi-level features, which are crucial for precise object classification and localization. Therefore, placing SPPF and CBAM after Stage 4 enables the refinement and fusion of multi-scale features, enhancing the network’s ability to handle complex driving scenarios with higher precision.

\noindent\textbf{Neck Branch.} The Neck module is composed of two key components. After the feature map passes through the backbone, it enters the FAFCENet, which consists of multiple FAFCE components and convolutional (Conv) layers. The FAFCE module is designed to improve feature fusion by ensuring the consistency and accuracy of multi-level features, while the convolutional layers further refine the features. By stacking multiple FAFCE components and applying convolutional layers, FAFCENet progressively refines feature information, enabling effective fusion of features across different levels. This fusion is crucial for handling the complex, multi-scale nature of autonomous driving environments. The PHFFNet then combines low-level and high-level features through the Hierarchical Fusion Block, addressing semantic discrepancies between non-adjacent layers and enhancing feature integration. The CASF mechanism in the blocks resolves conflicts between layers, ensuring the retention of critical information for accurate object detection. Finally, the refined feature map is passed into the Head layer for classification and localization. For details of the FAFCE and PHFFNet, refer to Sec. 3.3 and 3.4.

\noindent\textbf{Head Branch.} The Head module is responsible for receiving the feature map from the Neck and performing tasks such as object classification and bounding box regression. The Butter model uses four heads instead of the traditional three or more heads to balance multi-task processing and computational efficiency. The four heads allow for more detailed handling of multiple tasks in complex scenarios (such as object detection, lane line recognition, etc.), while avoiding the computational burden that comes with five or more heads. On the other hand, three heads may not fully meet the diverse demands of autonomous driving environments. Therefore, the use of four heads improves detection accuracy while maintaining good efficiency.

\label{subsection:BM}

\subsection{Frequency-Adaptive Feature Consistency Enhancement (FAFCE) Component}  

Current object detection models face challenges in hierarchical representation learning, mainly due to the lack of semantic information in low-level features and spatial loss in high-level features. Traditional fusion methods often cause information loss and inconsistencies, impairing boundary detection and small object recognition, which are vital in autonomous driving. Inspired by how shape priors enhance zero-shot segmentation~\cite{liu2023delving}, the FAFCE component employs frequency-aware damping and amplification to optimize the fusion of low-level and high-level features. This improves multi-scale information capture, enhancing detection accuracy and robustness in the dynamic, complex environments of autonomous driving.

We present the details of the FAFCE component, as illustrated in Fig.~\ref{fig:FAFCE} in the Supplementary Material, where the figure is included due to space constraints. FAFCE operates in 3 main stages: preliminary fusion, resampling and refined fusion. Before delving into the three stages, we will first provide an overview of traditional feature fusion methods in image processing, focusing on their formulations, which will set the stage for a detailed comparison with our proposed method.

A widely adopted approach to feature fusion can be expressed as:  
\begin{equation}
    \mathbf{B}^l = \mathbf{A}^l+ \mathcal{U}_{\theta}^{\mathrm{UP}}(\mathbf{B}^{l+1}),
\end{equation}  
where $\mathbf{A}^l \in \mathbb{R}^{2H \times 2W \times C}$ represents the $l$-th feature map produced by the backbone, and $\mathbf{B}^{l+1} \in \mathbb{R}^{H \times W \times C}$ denotes the fused feature at the $(l+1)$-th level. The operator $ \mathcal{U}_{\theta}^{\mathrm{UP}}$ refers to an upsampling function and $\theta$ is the learnable parameter. The two feature maps are assumed to have the same number of channels; otherwise, a $1 \times 1$ convolution can be applied as a projection function, which we omit here for simplicity. Despite its simplicity, this fusion approach causes semantic inconsistency and boundary misalignment. These issues harm dense prediction in autonomous driving. Simple interpolation spreads errors spatially, which leads to misclassifications. It also oversmooths outputs, weakening boundary localization. Additionally, the approach underutilizes boundary cues from lower-level features, which are vital for accurate detection in complex environments.

As shown in Figure~\ref{fig:FAFCE}, the proposed FAFCE consists of 2 important modules: the high-frequency amplifier, and the low-frequency damping. The 2 modules can be written as follows:

\noindent\textbf{High-Frequency Amplifier.} The amplifier amplifies high-frequency components of the input features, improving the fine details of the object boundaries. It is formulated as:
\begin{equation}
   \widetilde{\mathbf{A}}^{l} = \mathbf{A}^{l} + \mathcal{H}^{\mathrm{HF}}_{\phi} \left( W^{\mathrm{HF}}_{\mathrm{l}} \odot \mathbf{A}^{l} \right),
\end{equation}
where \( \mathcal{H}^{\mathrm{HF}}_{\phi} \) represents the high-frequency amplifier operation with learnable parameters \( \phi \), \( W^{\mathrm{HF}}_{\mathrm{l}} \) is the learnable filter matrix for high-frequency amplifier, and \( \odot \) denotes element-wise multiplication, enhancing the high-frequency features to provide better resolution for fine boundaries.

\noindent\textbf{Low-Frequency Damping:} The damping operation suppresses low-frequency components that may introduce noise into the feature map. The operation is formulated is
\begin{equation}
   \widetilde{\mathbf{B}}^{l+1} = \mathcal{U}_{\theta}^{\mathrm{UP}} \left( \mathcal{L}_{\psi}^{\mathrm{LF}} \left( \mathbf{B}^{l+1} \right) \right).
\end{equation}
The \( \mathcal{L}_{\psi}^{\mathrm{LF}} \) represents the low-frequency damping operation with learnable parameters \( \psi \), which helps suppress irrelevant low-frequency features. \( \mathcal{U}_{\theta}^{\mathrm{UP}} \) is the upsampling operator, restoring the spatial resolution of the feature map after damping. This operation ensures that the model focuses on the high-frequency details and ignores the irrelevant low-frequency components, leading to sharper and more accurate predictions.

\noindent\textbf{Three Stages of FAFCE.} These stages work sequentially to refine and enhance the feature maps in FAFCE, ultimately leading to the final fused features used for prediction.

\noindent\textbf{Stage 1: Preliminary Fusion}

In \textbf{Stage 1}, the initial fusion of features is performed using convolutional operations followed by high-frequency amplifier and low-frequency damping operations. The formula for this stage, including weight matrices, is as follows:
\begin{equation}
   \mathbf{B}_{Initial}^{l+1} = \mathbf{W}_B^{l+1} \odot \widetilde{\mathbf{B}}^{l+1} + \mathbf{W}_A^{l} \odot \widetilde{\mathbf{A}}^{l}.
\end{equation}
The \(\mathbf{W}_A^{l}\) and \(\mathbf{W}_B^{l+1}\) are the weight matrices associated with the feature \(\widetilde{\mathbf{A}}^l\) and \(\widetilde{\mathbf{B}}^{l+1}\).

\noindent\textbf{Stage 2: Resampling}

In \textbf{Stage 2}, the Feature Resampling Module fine-tunes the feature map by adjusting the spatial layout. Features are resampled to achieve better alignment across different layers, ensuring that high-level features are appropriately adjusted to match the finer details of the lower-level features. The formula for this stage is:
\begin{equation}
   \mathbf{B}_{Intermediate}^{l+1} = Re_{(u,v)} \left( \mathbf{B}_{Initial}^{l+1}, \widetilde{\mathbf{B}}^{l+1}\right),
\end{equation}
where \( Re_{(u,v)} \) is the resampling operation with displacement \((u,v)\), used to re-align the features from the previous stage, \( \mathbf{B}_{Initial}^{l+1} \) is the refined feature from the stage 1, and \( \widetilde{\mathbf{B}}^{l+1} \) represents the feature map from the amplifier. Stage 2 refines the spatial alignment of the features, ensuring that they are better aligned for the final fusion stage.

\noindent\textbf{Stage 3: Refined Fusion}

In \textbf{Stage 3}, the final fusion of the refined features is performed. This stage combines the results from the previous stages and applies further processing to ensure consistency and clarity of the final output. The formula for this stage is:
\begin{equation}
   \mathbf{B}^{l} = \mathbf{W}_B^{l+1} \odot \mathbf{B}_{Intermediate}^{l+1} + \mathbf{W}_A^{l} \odot \widetilde{\mathbf{A}}^{l}.
\end{equation}
Similarity, \(\mathbf{W}_A^{l}\) and \(\mathbf{W}_B^{l+1}\) are the weight matrices associated with the feature \(\widetilde{\mathbf{A}}^l\) and \(\mathbf{B}_{Intermediate}^{l+1}\). These weight matrices can be dynamically adjusted based on frequency correlations to ensure better fusion performance. After stage 3, the output leads to the fully refined feature map that can be used as the input to the PHFFNet module of the neck.

\label{subsection:FAFCE}
\subsection{Progressive Hierarchical Feature Fusion Network (PHFFNet)}  

After introducing FAFCE component, we also propose a novel paradigm PHFFNet for hierarchical feature learning, designed to enhance the object detection model's ability to learn and align features across different levels. PHFFNet progressively merges features from low to high levels, addressing the significant semantic disparity between non-adjacent layers, especially between the bottom and top features. This gap can weaken feature fusion in autonomous driving contexts, where precise object localization is crucial. By progressively narrowing this semantic gap, the PHFFNet architecture improves feature alignment, optimizing performance for detection tasks in complex driving environments.

\noindent\textbf{Mathematical Representation.}
We represent the progressive feature fusion process in matrix form, where each feature \( C_n \) and fused feature \( F_n \) are expressed as matrices or vectors.

\noindent\textbf{1. Initial Feature Fusion}

In the first step, we fuse the low-level features \( C_2 \) and \( C_3 \):
\begin{equation}
F_{23} = W_{2,3} \cdot \begin{bmatrix} C_2 \\ C_3 \end{bmatrix}.
\end{equation}
Here, \( W_{2,3} \) is the weight matrix used for the fusion of low-level features \( C_2 \) and \( C_3 \).

\noindent\textbf{2. Further Fusion}

Next, we fuse \( F_{23} \) with \( C_4 \), resulting in the new feature \( F_{234} \):
\begin{equation}
F_{234} = W_{2,3,4} \cdot \begin{bmatrix} F_{23} \\ C_4 \end{bmatrix},
\end{equation}
where \( W_{2,3,4} \) is the weight matrix that facilitates the fusion of \( F_{23} \) and \( C_4 \).

\noindent\textbf{3. Final Feature Fusion}

Finally, we fuse \( F_{234} \) with \( C_5 \), obtaining the final feature \( F_{2345} \):
\begin{equation}
F_{2345} = W_{2,3,4,5} \cdot \begin{bmatrix} F_{234} \\ C_5 \end{bmatrix}.
\end{equation}
Here, \( W_{2,3,4,5} \) is the weight matrix that facilitates the fusion of \( F_{234} \) and \( C_5 \).

We introduce the CASF mechanism within the Hierarchical Fusion Block to assign dynamic spatial weights to features at different levels during multi-level feature fusion. This approach strengthens the capacity of key layers to extract critical information while mitigating the impact of conflicting or irrelevant data from diverse objects and spatial regions, crucial for accurate detection in autonomous driving scenarios.

Let \( x_n^{\rightarrow l, ij} \) represents the feature vector at spatial position \( (i, j) \) from level \( n \) to level \( l \), and let \( y_l^{ij} \) denotes the resulting fused feature vector at position \( (i,j) \) and level \( l \). We integrate feature vectors from 3 distinct Hierarchical Fusion Block as illustrated in Fig.~\ref{fig:overview of architecture}. The fusion is performed by summing over these levels, with \( \lambda_n^{ij} \) representing the dynamic spatial weight for the feature vector at level \( n \) and spatial position \( (i,j) \). These weights dynamically adjust based on the significance of the features at each spatial location, allowing the fusion process to emphasize relevant features and suppress less important ones.

The fusion is mathematically expressed as:
\begin{equation}
y_l^{ij} = \sum_{n=1}^{3} \lambda_n^{ij} \cdot x_n^{\rightarrow l, ij},
\end{equation}
where \( \lambda_n^{ij} \) are the spatial fusion weights. The constraint
\begin{equation}
\sum_{n=1}^{3} \lambda_n^{ij} = 1 \quad \forall \, (i,j),
\end{equation}
ensures that the weights sum to unity at each spatial position, maintaining a balanced contribution from each feature level.

The \textbf{CASF} mechanism effectively resolves conflicts between features from different levels at the same spatial location, resulting in more robust feature representations. This significantly improves model performance in multi-scale object detection, crucial for autonomous driving, by leveraging the contextual relevance of features across varying levels.

\subsection{Loss Function}  

The total loss function \( \mathcal{L}_{\text{total}} \) is a weighted sum of three essential components: the bounding box regression loss \( \mathcal{L}_{\text{IoU}} \), the classification loss \( \mathcal{L}_{\text{cls}} \), and the distribution focal loss \( \mathcal{L}_{\text{dfl}} \). Each component is assigned a specific weight coefficient (\(\lambda_1\), \(\lambda_2\), \(\lambda_3\)), allowing for a balanced optimization during training. Specifically, the loss function is defined as:
\begin{equation}
\mathcal{L}_{\text{total}} = \lambda_1 \cdot \mathcal{L}_{\text{IoU}} + \lambda_2 \cdot \mathcal{L}_{\text{cls}} + \lambda_3 \cdot \mathcal{L}_{\text{dfl}}.
\end{equation}
\( \mathcal{L}_{\text{IoU}} \) quantifies the overlap between predicted and ground-truth bounding boxes, crucial for precise object localization. \( \mathcal{L}_{\text{cls}} \) ensures accurate object classification by employing focal loss or cross-entropy, which effectively addresses class imbalance, and is particularly critical for autonomous driving datasets. \( \mathcal{L}_{\text{dfl}} \) is designed to address class imbalance and enhance the model's focus on hard samples in object detection tasks.

This multi-task learning framework is central to optimizing hierarchical feature representation in autonomous driving, ensuring that both localization and classification are efficiently evaluated. The weighting coefficients (\(\lambda_1\), \(\lambda_2\), \(\lambda_3\)) fine-tune the model's ability to balance these critical tasks, allowing it to adapt to the specific challenges of real-time object detection in complex driving scenarios.



    
%

\label{subsection:loss}

\section{Experiments}  
\subsection{Experiment Settings}  
\noindent\textbf{Datasets and metrics.}  
We evaluate Butter on  three datasets specifically designed for autonomous driving—KITTI~\cite{geiger2012ready}, BDD100K~\cite{yu2020bdd100k} and Cityscapes~\cite{cityscapes}. The KITTI~\cite{geiger2012ready} dataset offers valuable in-vehicle perspective data across a variety of traffic scenarios, featuring over 15,000 labeled 2D images for object detection. The BDD100K~\cite{yu2020bdd100k} dataset contains annotations for a wide range of driving conditions and includes 100,000 frames of images, making it one of the largest driving video datasets available. The Cityscapes~\cite{cityscapes} dataset provides detailed semantic understanding of urban street scenes, consisting of 5,000 high-quality pixel-level annotated images from 50 cities, with 2,975 images for training, 500 for validation, and 1,525 for testing, covering 19 different categories. We evaluate our method with mAP@50 (mean Average Precision at IoU=0.5) to measure the precision of object detection at an IoU threshold of 0.5, and also consider the number of parameters to assess the model's lightweight nature and its ease of deployment. Additionally, we report GFlops (Giga Floating Point Operations per second) to assess the model's computational efficiency. 

\noindent\textbf{Implementation details.}  
We employ SGD as optimizer, applying weight decay to non-bias parameters, and train for 300 epochs. The input image size for all datasets is $640\times640$. During training, we use a batch size of 128,8,16 for BDD100K~\cite{yu2020bdd100k}, KITTI~\cite{geiger2012ready} and Cityscapes~\cite{cityscapes}, respectively. For inference, the batch size is set to 1 across all datasets. The Hyper-YOLO~\cite{hyperyolo}, YOLOv11~\cite{Khanam2024v11} and YOLOv12~\cite{Tian2025} are trained with the official code under the same settings as Butter. For fair comparison, we select variants with a comparable number of parameters to those of Butter. The selection of different scale versions under 15M of Hyper-YOLO~\cite{hyperyolo}, YOLOv11~\cite{Khanam2024v11}, and YOLOv12~\cite{Tian2025} as key components for our comparative experiments is due to the limited number of object detection models focused on autonomous driving in the past two years, with many older models being outdated. In contrast, the aforementioned three models have demonstrated remarkable performance in object detection tasks across all categories, including those in autonomous driving scenarios. Finally, the choice of 15M as the threshold is based on the need for lightweight models and ease of deployment in autonomous driving scenarios. 


\begin{table}[t]
\caption{The performance comparison between Butter and other methods on KITTI~\cite{geiger2012ready}.}
\label{tab:kitti}
\resizebox{\linewidth}{!}{
\begin{tabular}{@{}ccccc@{}}
\toprule
\textbf{Method} &  \textbf{Reference}&\textbf{mAP@50} & \textbf{GFlops} & \textbf{\# Params (M)} \\ \midrule
PIAENet512~\cite{piaenet}&  IS 24&83.4 & - & 9.1 \\
TOD-YOLOv7~\cite{tod-yolo}&  IS 25& \underline{93.2} & 102.6 & 83.5 \\
S-PANet~\cite{spanet}&  Intell. Veh. 24& 92.9 & 23.9& 6.8\\ \midrule
 Hyper YOLO-T~\cite{hyperyolo}& \multirow{3}{*}{TPAMI 24}
& 89.8& 8.9&3.0
\\
 Hyper YOLO-N~\cite{hyperyolo}& 
& 91.1& 10.8&3.9
\\
 Hyper YOLO-S~\cite{hyperyolo}& & 93.1 & 38.9&14.8\\ \midrule
 YOLOv11-N~\cite{Khanam2024v11}& \multirow{2}{*}{arXiv 24}& 88.7& 6.4&2.6\\
 YOLOv11-S~\cite{Khanam2024v11}& & 91.9& 21.3&9.4\\ \midrule
YOLOv12-N~\cite{Tian2025}&  \multirow{2}{*}{arXiv 25}&88.5&  \textbf{5.8} & \textbf{2.5} \\
YOLOv12-S~\cite{Tian2025}&   &90.3&  19.3&  9.1\\  \midrule
\rowcolor[HTML]{EFEFEF} Butter&  -&\textbf{94.4}& 31.0 & 5.4\\
\bottomrule
\end{tabular}
}
\end{table}

\begin{table}[t]
\caption{The performance comparison between Butter and other methods on BDD100K~\cite{yu2020bdd100k}.}
\label{tab:bdd-100k}
\resizebox{\linewidth}{!}{
\begin{tabular}{@{}ccccc@{}}
\toprule
\textbf{Method} &  \textbf{Reference}&\textbf{mAP@50} & \textbf{GFlops} & \textbf{\# Params (M)} \\ \midrule
MDNet~\cite{mdnet}&  PR 25&45.0& 12.3 & -\\
Yolop~\cite{yolop}&  Mach. Intell. 22&41.0& 8.49& - \\
OFFR-YOLO~\cite{offr-yolo}&  Syst. Appl. 24&45.3 & - & - \\ \midrule
 Hyper YOLO-T~\cite{hyperyolo}& \multirow{3}{*}{TPAMI 24}
& 46.7& 8.9&3.0\\
 Hyper YOLO-N~\cite{hyperyolo}& & 48.8& 10.8&
3.9\\
 Hyper YOLO-S~\cite{hyperyolo}& & \underline{53.2}& 38.9&
14.8\\ \midrule
 YOLOv11-N~\cite{Khanam2024v11}& \multirow{2}{*}{arXiv 24}& 44.5& 6.3&
2.6\\
 YOLOv11-S~\cite{Khanam2024v11}& & 50.6& 21.3&9.4\\\midrule
YOLOv12-N~\cite{Tian2025}&   \multirow{2}{*}{arXiv 25}&44.2&  \textbf{5.8}&  
\textbf{2.5}\\
YOLOv12-S~\cite{Tian2025}&  & 52.1&  19.3& 9.1\\
\rowcolor[HTML]{EFEFEF} Butter&  -&\textbf{53.7}& 30.9 & 5.4\\ 
\bottomrule
\end{tabular}
}
\end{table}

\begin{table}[t]
\caption{The performance comparison between Butter and other methods on Cityscapes~\cite{cityscapes}.}
\label{tab:cityscapes}
\resizebox{\linewidth}{!}{
\begin{tabular}{@{}ccccc@{}}
\toprule
\textbf{Method} &  \textbf{Reference}
&\textbf{mAP@50} & \textbf{GFlops} & \textbf{\# Params (M)} \\ \midrule
 Yolop~\cite{yolop}& Mach. Intell. 22& 20.3& 8.49&-\\ 
MDNet&  PR 25
&37& 12.3& -\\ \midrule
 Hyper YOLO-T~\cite{hyperyolo}& \multirow{3}{*}{TPAMI 24}& 45.1& 8.9&3.0
\\
 Hyper YOLO-N~\cite{hyperyolo}& & 48.4& 10.8&3.9
\\ 
Hyper YOLO-S~\cite{hyperyolo}&  &\underline{51.6}& 38.9& 14.8\\ \midrule
 YOLOv11-N~\cite{Khanam2024v11}& \multirow{2}{*}{arXiv 24}& 42.8& 6.3&
2.6\\
 YOLOv11-S~\cite{Khanam2024v11}& & 49.8& 21.3&9.4\\ \midrule
 YOLOv12-N~\cite{Tian2025}& \multirow{2}{*}{arXiv 25}& 45.0& \textbf{5.8}&
\textbf{2.5}
\\ 
YOLOv12-S~\cite{Tian2025}&   &51.4&  19.3&  9.1\\  \midrule
\rowcolor[HTML]{EFEFEF} Butter&  -&\textbf{53.2}& 31.1& 5.4\\
\bottomrule
\end{tabular}
}
\end{table}
\subsection{Main Results}  
The object detection results on the KITTI~\cite{geiger2012ready}, BDD100K~\cite{yu2020bdd100k}, and Cityscapes~\cite{cityscapes} datasets are presented in Tab.~\ref{tab:kitti}, Tab.~\ref{tab:bdd-100k}, Tab.~\ref{tab:cityscapes}. On the KITTI~\cite{geiger2012ready} dataset, Butter's mAP@50 exceeds that of the existing SOTA method, TOD-YOLOv7~\cite{tod-yolo}, by 1.2, while its GFlops are only approximately one-third of TOD-YOLOv7~\cite{tod-yolo}'s. On both the BDD100K~\cite{yu2020bdd100k} and Cityscapes~\cite{cityscapes} datasets, Butter exhibits superior performance relative to the parameter-efficient variant of the current state-of-the-art method, Hyper-YOLO-S~\cite{hyperyolo}. It achieves higher mAP@50, with a particularly notable improvement of 1.6 mAP@50 on Cityscapes~\cite{cityscapes}. Overall, the total parameter count is reduced by approximately 64\% compared to Hyper-YOLO-S~\cite{hyperyolo}. Remarkably, Butter consistently outperforms other prominent real-time detection models, including Hyper-YOLO~\cite{hyperyolo}, YOLOv11~\cite{Khanam2024v11}, and YOLOv12~\cite{Tian2025}, across all three datasets. These results demonstrate that Butter achieves the optimal trade-off between parameter efficiency, deployability, and detection accuracy.

\subsection{Ablation and Analyses}
\begin{table}[t]
\caption{Ablation study on the proposed methods, with all models trained from scratch on the KITTI~\cite{geiger2012ready}. }
\label{tab:ablation}
\resizebox{\linewidth}{!}{
\begin{tabular}{@{}c|cccccc@{}}
\toprule
\textbf{Model} &  \textbf{Backbone}&\textbf{Head}& \textbf{PHFFNet} & \textbf{FAFCE} & \textbf{mAP@50} & \textbf{\# Params} \\ \midrule
(1) &   HGNetV2
&
3&  &  & 92.3 & 9.7 \\
(2) &  HGNetV2$^{\textrm{light}}$&
3&  &  & 92.2 & 6.8 \\
(3) &  HGNetV2$^{\textrm{light}}$&4&  &  & 93.9 & 9.8 \\
(4) &  HGNetV2$^{\textrm{light}}$&4& \checkmark &  & 93.2 & 6.9 \\
\rowcolor[HTML]{EFEFEF} 
Butter&  HGNetV2$^{\textrm{light}}$&4& \checkmark & \checkmark & \textbf{94.4} & \textbf{5.4} \\ \bottomrule
\end{tabular}
}
\end{table}
\noindent\textbf{Ablation on proposed methods.} To investigate the
 impact of our proposed methods, we conduct an ablation
 study on gradually increasing components by using KITTI~\cite{geiger2012ready} datasets. The results are presented in Tab.~\ref{tab:ablation}.  Comparing row (1) and row (2), we replace the origininal HGNetV2 with its lightweight version, resulting in a nearly lossless reduction of 2.9M parameters. In row (3), we change the traditional 3-head configuration to a 4-head configuration. This enables Butter to handle multiple tasks in complex scenarios while maintaining a balanced trade-off between accuracy and parameter-efficiency. However, in the neck branch, it directly concatenates or cascades features from different layers, which requires more parameters to handle these larger feature maps. Therefore, in row (4), our PHFFNet progressively combines low-level and high-level features, allowing the model to capture sufficient contextual and abstract information with fewer parameters. This results in a 2.9M parameter reduction compared to row (3), with only a slight performance drop. Finally, we add the FAFCE module before PHFFNet to enhance feature consistency during the feature map representation learning process. Comparing row (4) and Butter, after adding FAFCE, it is encouraging that the parameter count was reduced by 1.5M, while the mAP@50 still increased by 1.2.
 
\begin{figure}[t]
\begin{center}
   \includegraphics[width=1\columnwidth]{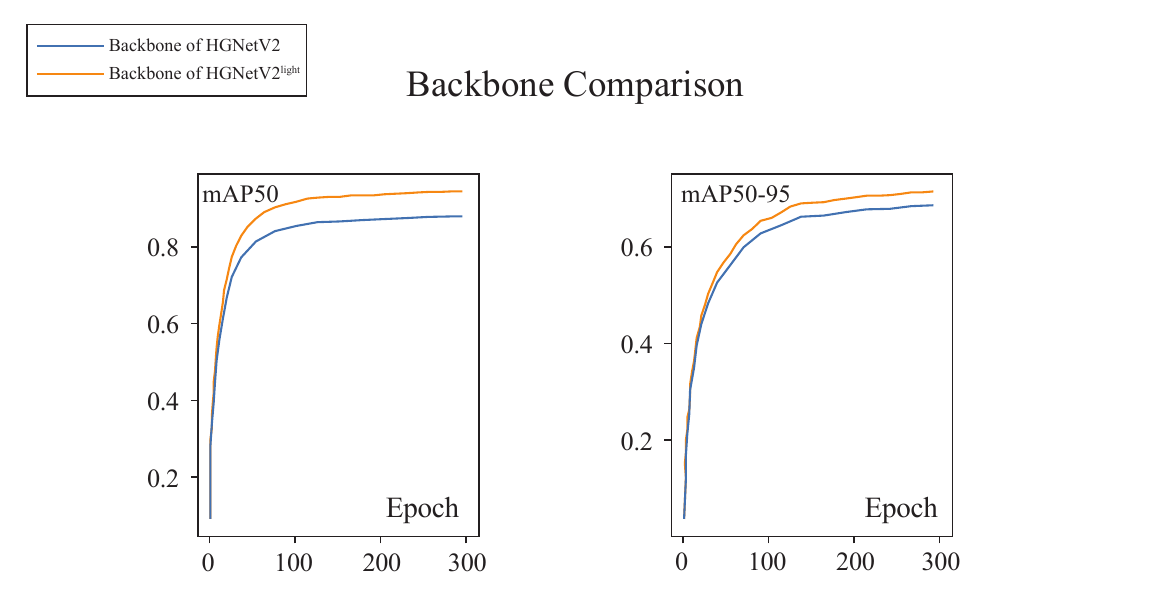}
\end{center}

\caption{Comparison of model performance with different backbones on the KITTI~\cite{geiger2012ready} dataset. The left plot shows the mAP50, and the right plot shows the mAP50-95.}
\label{fig:Training}
\vspace{-0.5cm}

\end{figure}

\noindent\textbf{Effect of the Butter backbone.} In this experiment, we replace the backbone of the Butter model with the original HGNetV2. All models were trained for 300 epochs on the KITTI~\cite{geiger2012ready} dataset. As shown in Fig.~\ref{fig:Training}, under the same number of training epochs, the Butter model with the lightweight backbone consistently achieves higher mAP50 and mAP50-95 scores. Moreover, based on the overall trend and curve fitting, the model with the lightweight backbone also demonstrates better convergence behavior.

\begin{figure}
    \centering
    \includegraphics[width=\linewidth]{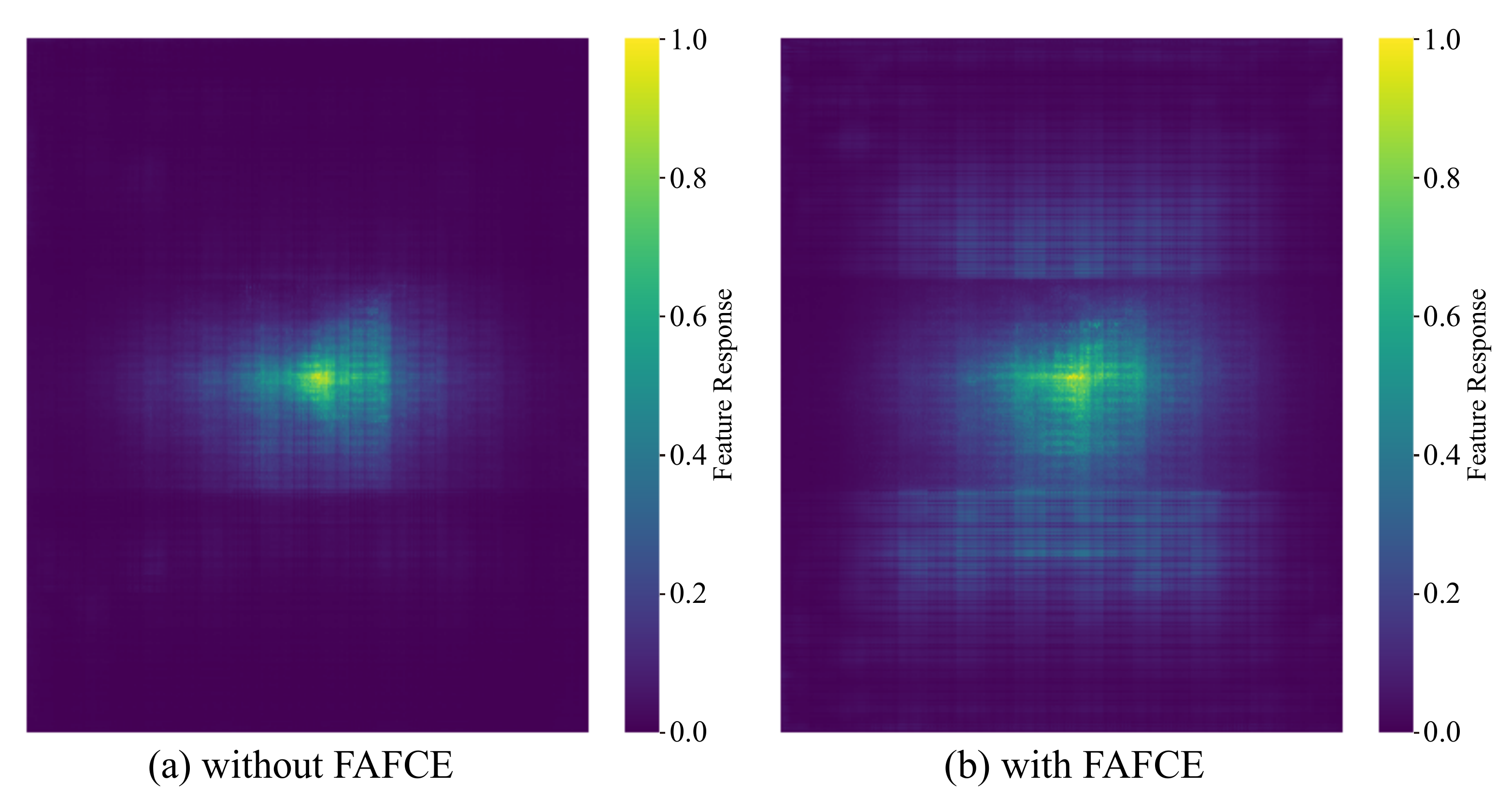}
    \caption{Feature Response in Receptive Field.}
    \label{fig:Feature_Response}
    \vspace{-10pt} 
\end{figure}

\begin{figure}
    \centering
    \includegraphics[width=\linewidth]{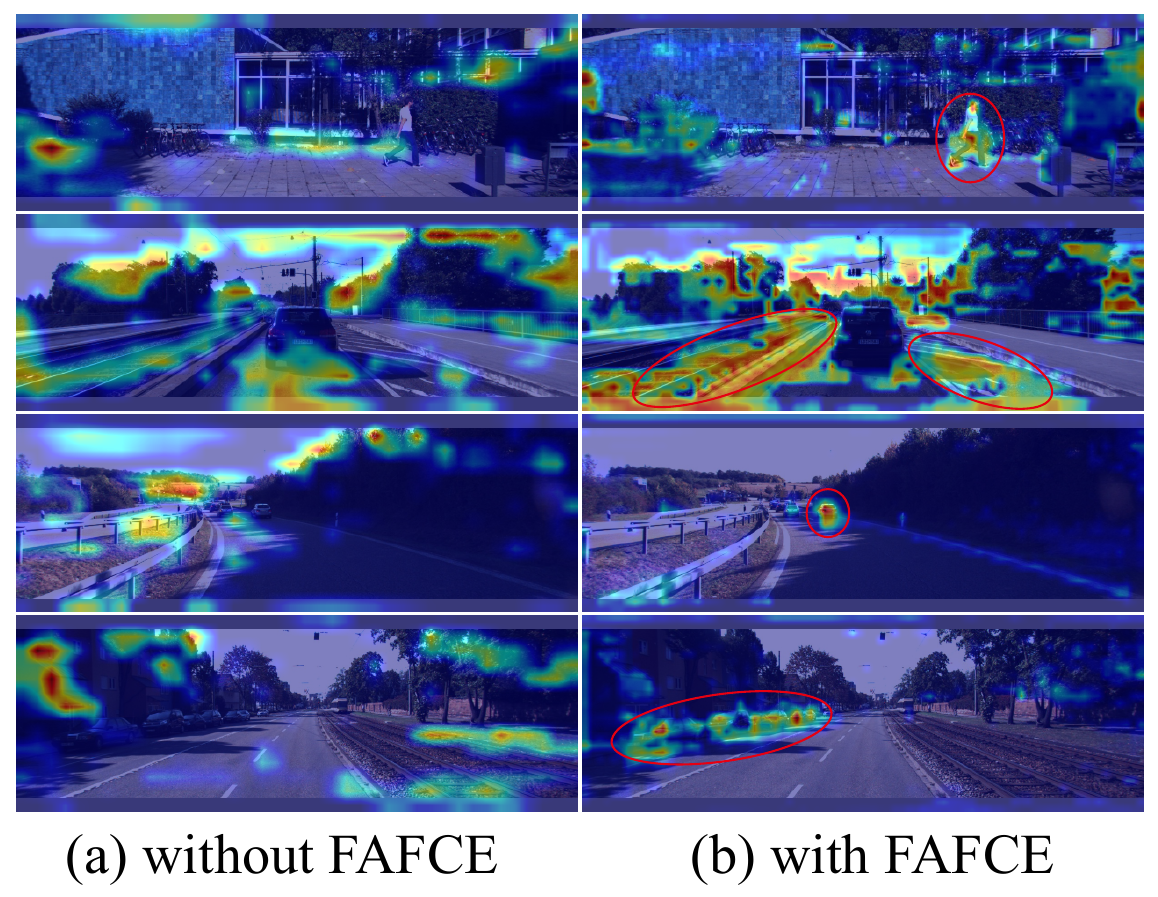}
    \caption{Heatmap Comparison of Model Attention in Butter.}
    \label{fig:Heatmaps}
    \vspace{-10pt} 
\end{figure}

\noindent\textbf{Effect of the FAFCE.} 
As shown in Fig.~\ref{fig:Feature_Response}, the receptive field after applying FAFCE component demonstrates a significant enhancement in feature response, with more pronounced color variations, reflecting the model's increased attention to the image's context and details.
In Fig.~\ref{fig:Heatmaps}, we compare the heatmap of model's attention both with and without the FAFCE module in KITTI~\cite{geiger2012ready} dataset. (a) Without FAFCE, the attention is less focused and more scattered, leading to imprecise localization of the target objects. (b) With FAFCE, the attention is significantly improved, showing a more concentrated focus on the objects and their surrounding context. This demonstrates the enhanced detection performance and better contextual understanding that FAFCE provides, highlighting its effectiveness in improving object detection accuracy in autonomous driving.

\section{Conclusion}

In this paper, we propose Butter, a novel model designed for efficient object detection in autonomous driving scenarios. The core innovation lies in the Neck architecture, specifically through the introduction of two key components: the FAFCE component and the PHFFNet. These innovations enhance the consistency and precision of multi-scale features, enabling robust feature fusion across different levels and improving object detection performance and parameter-efficiency without a heavy computational burden. The model achieves a trade-off between deployability, accuracy, and computational efficiency, making it highly suitable for object detection tasks in autonomous driving scenarios. In future research, we will apply hierarchical representation learning to video-based models, aiming to develop object detection systems better suited for real-world autonomous driving. We also plan to extend our model's principles to tasks like semantic segmentation and general object detection, further validating its versatility across diverse applications.



\newpage
\bibliographystyle{ACM-Reference-Format}
\bibliography{sample-base}

\newpage
\appendix
\twocolumn[
\begin{@twocolumnfalse}
\begin{center}
    \Large \textbf{Butter: Frequency Consistency and Hierarchical Fusion for Autonomous Driving Object Detection}
\end{center}
\begin{center}
    \large \textbf{Supplementary Material}
\end{center}
\end{@twocolumnfalse}
]

\begin{figure*}[!htp]
   \centering
   \includegraphics[width=1\textwidth]{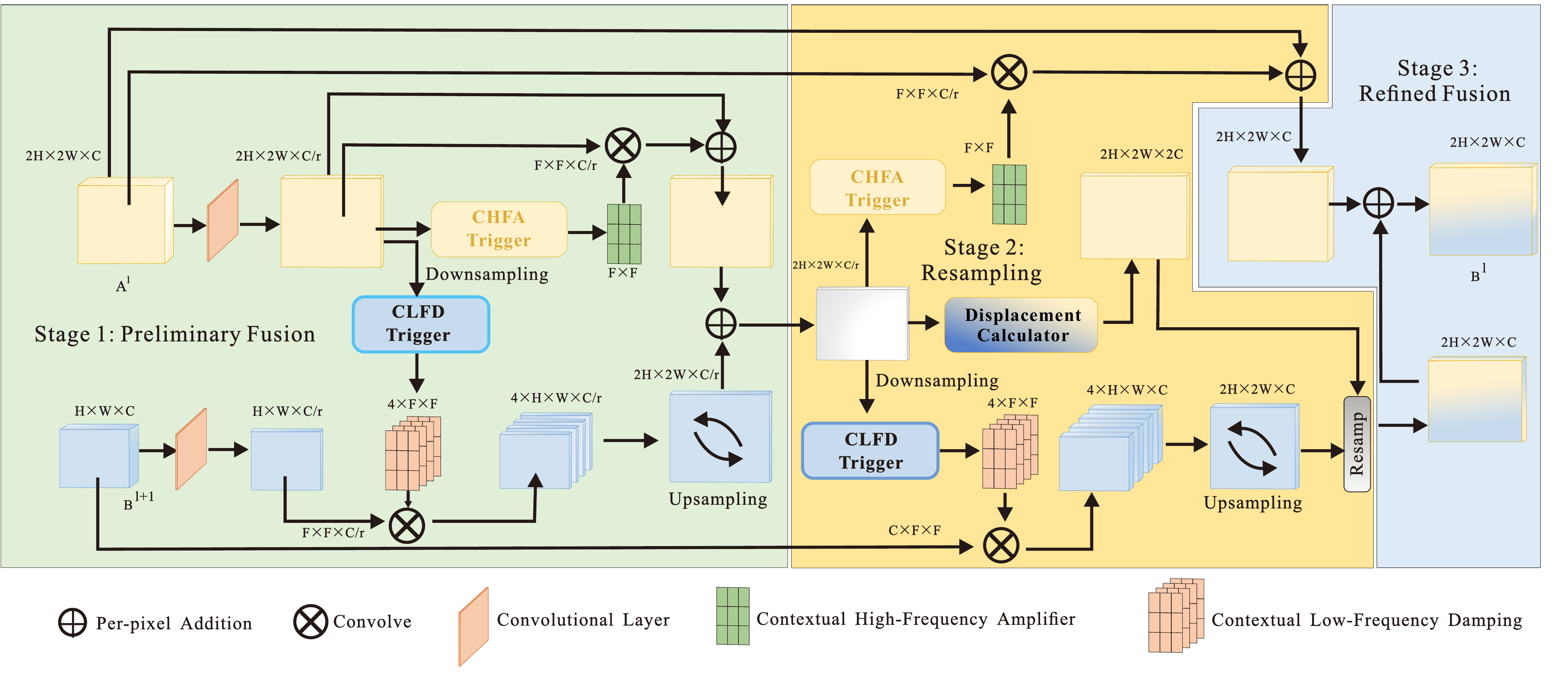}
   \vspace{-18pt}
\setlength{\abovecaptionskip}{0.5cm}
   \caption{
Architecture of the FAFCE component, which operates in three stages. (a) Stage 1: Preliminary fusion of features with pixel downsampling, followed by CLFD and CHFA triggers, and final per-pixel addition of features. (b) Stage 2: The process is primarily divided into two branches. One branch undergoes pixel downsampling, followed by the CLFD trigger, pixel upsampling, and integration with the output from the displacement calculator, followed by a resampling step before entering Stage 3. The other branch passes through the CHFA trigger, then undergoes convolution and per-pixel addition before directly entering Stage 3.(c) Stage 3: Final refined fusion of low and high-level features for enhanced output.
}
\label{fig:FAFCE}
\end{figure*}
\section{Implementation Details}

\subsection{Hyper-parameters}
The main hyper-parameters of Butter are listed in Tab.~\ref{tab:hyper-params}

\begin{table}[htbp]
    \centering
    \begin{tabular}{c|c} 
    \toprule
         \textbf{Item}&  \textbf{Value}\\ \midrule
 optimizer&SGD\\
 learning rate&1e-2\\
 weight decay&5e-4 for non-bias\\
 momentum &0.937\\
 epochs&300\\
 input size&$640\times 640$\\
 bbox loss weight&7.5\\
 cls loss weight&0.5\\ 
         dfl loss weight&  1.5\\ 
    \bottomrule
    \end{tabular}
    \caption{Main hyper-parameters of Butter. }
    \label{tab:hyper-params}
\end{table}

\subsection{FAFCE Architecture}

Fig.~\ref{fig:FAFCE} presents the detailed architecture of our proposed FAFCE module, which could not be included in the main text due to space limitations. As introduced in the main manuscript, FAFCE is designed to address the limitations of traditional feature fusion strategies by incorporating a three-stage process: preliminary fusion, resampling, and refined fusion. Each stage is carefully constructed to enhance the complementary strengths of multi-source features while minimizing redundancy. The figure provides a comprehensive visual representation of the data flow and module interactions across these stages, offering a clearer understanding of how FAFCE achieves its superior fusion capability compared to conventional approaches.

\subsection{Contextual Low-Frequency Damping (CLFD) Trigger}

The Contextual Low-Frequency Damping (CLFD) Trigger, as illustrated in (Fig.~\ref{fig:overview2} (a)), is designed to generate flexible low-frequency damping, which smooths out high-level features to reduce inconsistencies and facilitates the subsequent upsampling of these features. To achieve effective low-frequency damping, it is essential to integrate both high- and low-level features. Consequently, the CLFD Trigger processes the initially fused \( M^l \) and generates spatially varying low-frequency dampings. It uses a \( 3 \times 3 \) convolutional layer, followed by a softmax layer, as represented by the following equation

\begin{equation}
\overline{N}^l = \text{Conv}_{3 \times 3}(M^l),
\end{equation}
and
\begin{equation}
\overline{Q}^{l,a,b}_{i,j} = \text{Softmax}(\overline{N}^l_{i,j}) = \frac{\exp(\overline{N}^{l,a,b}_{i,j})}{\sum_{a,b \in \Omega} \exp(\overline{N}^{l,a,b}_{i,j})}.
\end{equation}
The \( \overline{N}^l \in \mathbb{R}^{F^2 \times 2H \times 2W} \) represents the spatially varying damping weights, where \( F \) denotes the kernel size for the damping operation. After reshaping, \( \overline{N}^l \) holds \( F \times F \) damping values for each spatial location. In this case, \( \Omega \) refers to the size of \( F \times F \). Following the application of a kernel-wise softmax, which ensures that all damping values are positive and their sum equals one, the resulting output is smooth, and the damping is represented as \( \overline{Q} \in \mathbb{R}^{F^2 \times 2H \times 2W} \).
Following the application of a kernel-wise softmax, the damping values are ensured to be positive and their sum equals one. The resulting output is smooth, and the damping is represented as \( \overline{Q} \in \mathbb{R}^{F^2 \times 2H \times 2W} \).

Then, we need to upscale \( B^{l+1} \in \mathbb{R}^{C \times H \times W} \). The first step involves reshaping \( \overline{Q}^l \) by downsampling, which reduces the height and width by half while increasing the channel dimension by a factor of 4. We then divide the channels into 4 distinct groups, with each group containing a spatially varying low-frequency damping, represented as \( \overline{Q}^l_g \in \mathbb{R}^{F^2 \times H \times W} \), where \( g \in \{1, 2, 3, 4\} \) denotes the group number. As a result, we obtain 4 separate groups of low-frequency damped features, denoted as \( B^{l+1,g} \in \mathbb{R}^{C \times H \times W} \), which are then rearranged as\( B^{l+1} \in \mathbb{R}^{C \times 2H \times 2W} \) and upsampled by a factor of 2, yielding the final feature as:

\begin{equation}
{B}^{l+1,g}_{i,j} = \sum_{a,b \in \Omega} \overline{Q}^{l,g,a,b}_{i,j} \cdot B^{l+1}_{i+a,j+b}, 
\end{equation}
and
\begin{equation}
{B}^{l+1} = \mathcal{U}_{\theta}^{\mathrm{UP}}(\tilde{B}^{l+1,1}, \tilde{B}^{l+1,2}, \tilde{B}^{l+1,3}, \tilde{B}^{l+1,4}).
\end{equation}
The operator $ \mathcal{U}_{\theta}^{\mathrm{UP}}$ which we have mentioned in the main text refers to an upsampling function and $\theta$ is the learnable parameter.

\subsection{Displacement Calculator}

Although the CLFD trigger improves intra-category similarity by smoothing features, it may struggle to correct large regions of inconsistent features or refine narrow and boundary areas. Enlarging the low-frequency damping size helps to address larger inconsistent regions but negatively affect the sharpness of thin and boundary regions. On the other hand, reducing the damping size helps to maintain the integrity of thin and boundary areas but may limit its ability to correct widespread feature inconsistencies.

\begin{figure*}[!htp]
   \centering
   \includegraphics[width=1\textwidth]{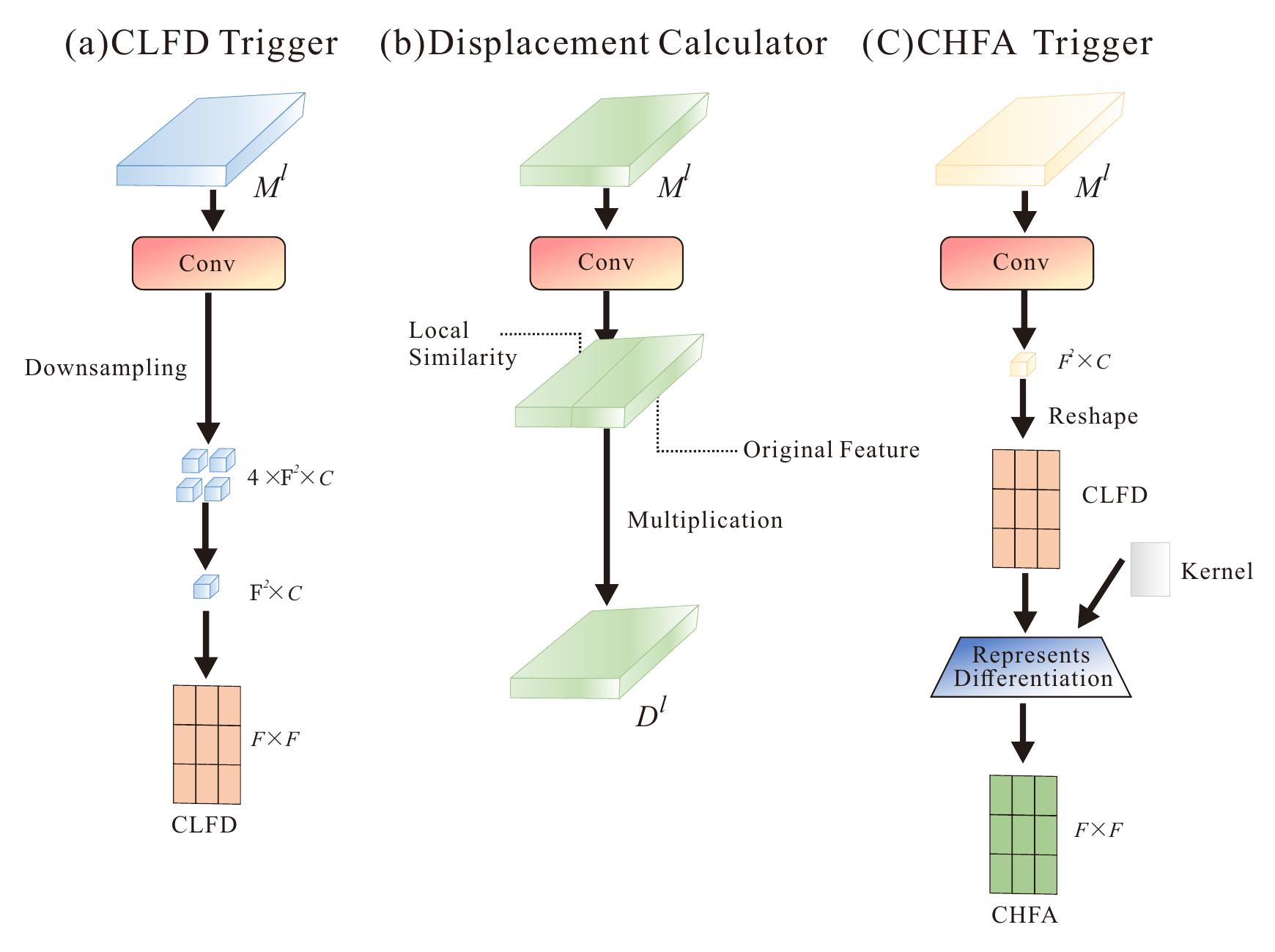}
\setlength{\abovecaptionskip}{0.5cm}
   \caption{Process of 3 Key Modules in the FAFCE Component.
(a) Contextual Low-Frequency Damping (CLFD) Trigger
(b) Displacement Calculator
(c) Contextual High-Frequency Amplifier (CHFA) Trigger
}

\label{fig:overview2}
\end{figure*}

To resolve this issue, we introduce the displacement calculator, as illustrated in (Fig.~\ref{fig:overview2} (b)). The idea behind this is based on the observation that neighboring features with low intra-category similarity often correspond to features with high intra-category similarity. Therefore, the displacement calculator starts by calculating the local cosine similarity:

\begin{equation}
S^{l,a,b}_{i,j} = \frac{\sum_{c=1}^{C} M^l_{c,i,j} \cdot M^l_{c,i+a,j+b}}{\sqrt{\sum_{c=1}^{C} (M^l_{c,i,j})^2} \cdot \sqrt{\sum_{c=1}^{C} (M^l_{c,i+a,j+b})^2}},
\end{equation}
where \( S \in \mathbb{R}^{8 \times H \times W} \) represents the local cosine similarity between each pixel and its 8 neighboring pixels. This process encourages the displacement calculator to focus on features with high intra-category similarity, effectively reducing the ambiguity in boundary areas or regions with inconsistent intra-category features.

To enhance the ability to refine inconsistent regions and preserve fine structures, the displacement calculator is employed to predict spatial offsets for feature resampling. It takes both the fused feature map \( M^l \) and the local similarity map \( S^l \) as inputs. These inputs are concatenated and passed through two separate \( 3 \times 3 \) convolutional layers: one to estimate the displacement orientation and the other to determine its scale. The outputs are defined as:

\begin{equation}
D^l = O^l \cdot P^l,
\end{equation}
\begin{equation}
O^l = \text{Conv}_{3 \times 3}(\text{Concat}(M^l, S^l)),
\end{equation}
and
\begin{equation}
P^l = \text{Sigmoid}(\text{Conv}_{3 \times 3}(\text{Concat}(M^l, S^l))).
\end{equation}
Here, \( O^l \) encodes the orientation of displacement, while \( P^l \) regulates its intensity through a sigmoid activation. The resulting displacement field \( D^l \) guides the model to shift high-level features toward areas with higher intra-class consistency. This targeted resampling mechanism effectively sharpens object boundaries and improves recognition in regions with spatial inconsistencies.

\subsection{Contextual High-Frequency Amplifier (CHFA) Trigger}

While the CLFD Trigger and displacement calculator are effective in recovering high-level features with refined boundaries and high intra-class consistency during upsampling, they cannot fully restore the fine boundary details. These details are lost in lower-level features due to downsampling.

This limitation is rooted in the Nyquist-Shannon Sampling Theorem, which states that frequencies above the Nyquist frequency, defined as half the sampling rate, are irretrievably lost during downsampling. For instance, if the high-level feature is downsampled by a factor of 2 relative to the low-level feature (such as through a \(1 \times 1\) convolution with a stride of 2), the sampling rate is reduced to \(\frac{1}{2}\). As a result, any frequencies exceeding \(\frac{1}{4}\) of the original frequency are subject to aliasing, meaning that they cannot be recovered during the upsampling process.

To clarify, we first transform the feature map \(\mathbf{A} \in \mathbb{R}^{C \times H \times W}\) into the frequency domain using the Discrete Fourier Transform (DFT), denoted as \(\mathbf{A}_F = \mathcal{F}(\mathbf{A})\), which is computed as:

\begin{equation}
\mathbf{A}_F(u,v) = \frac{1}{HW} \sum_{h=0}^{H-1} \sum_{w=0}^{W-1} \mathbf{A}(h,w)e^{-2\pi j(uh+vw)},
\end{equation}
where \(\mathbf{A}_F \in \mathbb{C}^{C \times H \times W}\) is the resulting complex-valued array after applying the DFT. Here, \(H\) and \(W\) represent the height and width of the feature map, and \(h, w\) are the coordinates of \(\mathbf{A}\). The frequencies in the height and width dimensions are normalized by \(|u|\) and \(|v|\). 
Following the Nyquist-Shannon Sampling Theorem, any high-frequency components that exceed the Nyquist frequency are lost during downsampling. Specifically, frequencies with \(|u| > \frac{1}{4}\) or \(|v| > \frac{1}{4}\) belong to the set \(\mathcal{H}^+ = \{(u,v) \mid |u| > \frac{1}{4} \text{ or } |v| > \frac{1}{4}\}\), and these frequencies are permanently aliased and cannot be recovered in the downsampled high-level features.

To overcome this issue, we utilize the CHFA trigger to recover the boundary details that are typically lost during downsampling. The CHFA trigger takes the initially fused feature map \(\mathbf{M}^l\) as input and generate spatially-variant high-frequent amplifier. It consists of a \(3 \times 3\) convolutional layer, followed by a CLFD (softmax layer) and a kernel represents differentiation operation, as shown in (Fig.~\ref{fig:overview2} (c)). These steps are defined as:
\begin{equation}
\widehat{\mathbf{V}}^l = \text{Conv}_{3 \times 3}(\mathbf{M}^l),
\end{equation}
and
\begin{equation}
\widehat{\mathbf{W}}^{l,p,q}_{i,j} = \mathbf{E} - \text{Softmax}(\widehat{\mathbf{V}}^l_{i,j}) = \mathbf{E}^{p,q} - \frac{\exp(\widehat{\mathbf{V}}^{l,p,q}_{i,j})}{\sum_{p,q \in \Omega} \exp(\widehat{\mathbf{V}}^{l,p,q}_{i,j})}.
\end{equation}
In this, \(\widehat{\mathbf{V}}^l \in \mathbb{R}^{\hat{F}^2 \times H \times W}\) represents the initial kernel values at each location \((i,j)\), where \(\hat{F}\) denotes the kernel size of the high-frequency amplifier. To ensure that the final kernels \(\widehat{\mathbf{W}}^l\) are high-frequency, we first obtain low-frequency kernels using kernel-wise softmax of CLFD and then invert them by subtracting from the identity kernel \(\mathbf{E}\), which has the values \([[0,0,0], [0,1,0], [0,0,0]]\) when \(\hat{F} = 3\). 

After applying the high-frequency amplifier and adding them residually, we obtain the enhanced feature map, expressed as:

\begin{equation}
\widehat{\mathbf{A}}^l_{i,j} = \mathbf{A}^l_{i,j} + \sum_{p,q \in \Omega} \widehat{\mathbf{W}}^{l,p,q}_{i,j} \cdot \mathbf{A}^l_{i+p,j+q}.
\end{equation}

\subsection{Extra Ablation Study}

To better understand the contributions of each module beyond the main text experiments, we conducted additional ablation studies on the KITTI~\cite{geiger2012ready} dataset. These include component-wise ablation of the FAFCE modules, detection head variants, and targeted analysis on small object detection. The results offer deeper insight into the internal structure and performance-efficiency trade-offs of our model.

\paragraph{(a) Component-wise Ablation of FAFCE on KITTI~\cite{geiger2012ready}.} 
We isolate \textbf{CLFD} and \textbf{CHFA} to assess their individual and joint effects within the full FAFCE structure.

\begin{table}[h]
\centering
\caption{Component-wise ablation of CLFD and CHFA modules in FAFCE (KITTI~\cite{geiger2012ready} Dataset)}
\begin{tabular}{lcc}
\toprule
\textbf{Configuration} & \textbf{mAP@50} & \textbf{\# Params (M)} \\
\midrule
No FAFCE      & 93.2 & 6.9 \\
CLFD only     & 93.8 & 6.3 \\
CHFA only   & 94.0 & 6.4 \\
CLFD + CHFA (FAFCE)     & \textbf{94.4} & \textbf{5.4} \\
\bottomrule
\end{tabular}
\end{table}

These results confirm the \textit{complementary nature} of CLFD and CHFA and show that FAFCE achieves the best accuracy--efficiency trade-off. The reduction in parameter count is due to the replacement of heavier fusion blocks with our lightweight CLFD/CHFA designs.

\paragraph{(b) Detection Head Comparison on KITTI~\cite{geiger2012ready}.}
We also compare detection head configurations to study the balance between performance and complexity.

\begin{table}[h]
\centering
\caption{Ablation of detection head variants in the Head Branch (KITTI~\cite{geiger2012ready} Dataset)}
\begin{tabular}{ccc}
\toprule
\textbf{Head Count} & \textbf{mAP@50} & \textbf{\# Params (M)} \\
\midrule
3 & 92.2 & \textbf{6.8} \\
4 & \textbf{93.9} & 9.8 \\
5 & 93.4 & 13.1 \\
\bottomrule
\end{tabular}
\end{table}

The 4-head variant delivers the best trade-off: fewer heads reduce capacity, while more increase overhead without clear gains. This balance reinforces our design decisions in maintaining architectural efficiency without compromising detection quality.

\paragraph{(c) Small Object Detection Capability.}
Building upon the prior experiments, we further investigate the model's effectiveness on small object detection—an area where accurate boundary localization and fine-grained feature retention are critical. In response to reviewer WVLL's suggestion, we report the average precision (AP) on small objects under the  MS COCO~\cite{lin2014microsoft} definition (i.e., object area < 32$^2$ pixels), using the KITTI~\cite{geiger2012ready} dataset. The results are summarized in Table~\ref{tab:small-object}.

\begin{table}[h]
\centering
\caption{Ablation Study Results for FAFCE Module on Small Object AP (KITTI~\cite{geiger2012ready} Dataset)}
\label{tab:small-object}
\begin{tabular}{lcc}
\toprule
\textbf{Configuration} & \textbf{AP\textsubscript{small}@[0.50:0.95]} & \textbf{\# Params (M)} \\
\midrule
No FAFCE  & 41.8 & 6.9 \\
Include FAFCE  & \textbf{43.9} & \textbf{5.4} \\
\bottomrule
\end{tabular}
\end{table}

The inclusion of FAFCE significantly improves small object AP, confirming its advantage in boundary-aware fusion. These findings will be included in the camera-ready version and support the broader claim that our method is particularly effective in challenging fine-scale detection scenarios.

\subsection{Loss Function}
We design a multi-component loss function that explicitly optimizes for localization accuracy, classification confidence and distribution-aware regression. 

The bounding box regression loss $\mathcal{L}_{\text{IoU}} $ optimizes the discrepancy between predicted bounding boxes and ground truth boxes. The mathematical formulation is as follows:

\begin{equation}
\begin{aligned}
\mathcal{L}_{\text{IoU}} = & \lambda_{\text{coord}} \sum_{i=0}^{S^2} \sum_{j=0}^{B} l_{ij}^{\text{obj}} \left[ (x_i - \hat{x}_i)^2 + (y_i - \hat{y}_i)^2 \right] \\
& + \lambda_{\text{coord}} \sum_{i=0}^{S^2} \sum_{j=0}^{B} l_{ij}^{\text{obj}} \left[ (\sqrt{w_i} - \sqrt{\hat{w}_i})^2 + (\sqrt{h_i} - \sqrt{\hat{h}_i})^2 \right].
\end{aligned}
\end{equation}
\( S \) is Grid size, \( B \) is the number of bounding boxes predicted per grid cell, \( l_{ij}^{\text{obj}} \) is Indicator function (1 if the \( j \)-th bounding box in the \( i \)-th grid cell is responsible for detecting an object; 0 otherwise). \( x_i, y_i \) is the predicted bounding box center coordinates. \( \hat{x}_i, \hat{y}_i \) is ground truth bounding box center coordinates. \( w_i, h_i \) is predicted bounding box width and height. \( \hat{w}_i, \hat{h}_i \) is ground truth bounding box width and height. \( \lambda_{\text{coord}} \) is Weighting coefficient to balance the contribution of coordinate loss. 

\( \mathcal{L}_{\text{cls}} \) ensures accurate object classification by employing focal loss or cross-entropy, which is expressed as: 
\begin{equation}
\begin{aligned}
 \mathcal{L}_{\text{cls}} = \sum_{i=0}^{S^2} l_i^{obj} \sum_{c \in \text{classes}} (p_i(c) - \hat{p}_i(c))^2.
 \end{aligned}
\end{equation}
\( S \) is the size of the grid. \( l_i^{obj} \) indicates whether the \( i \)-th grid cell contains an object. \( p_i(c) \) is the model's predicted probability that the object in the \( i \)-th grid cell belongs to class \( c \). \( \hat{p}_i(c) \) is the ground truth label, representing whether the object in the \( i \)-th grid cell actually belongs to class \( c \).  

$\mathcal{L}_{\text{dfl}}$ is designed to address class imbalance by down-weighting easy samples and focusing on hard samples.
\begin{equation}
\begin{aligned}
\mathcal{L}_{\text{dfl}} = \sum_{i=1}^N \sum_{c=1}^C y_{ic} \left( \alpha (1 - p_{ic})^\gamma \log(p_{ic}) + (1 - \alpha) p_{ic}^\gamma \log(1 - p_{ic}) \right).
 \end{aligned}
\end{equation}
\( N \) is the number of samples, and \( C \) is the number of classes. \( y_{ic} \) is the ground-truth label (one-hot encoded, with only one element being 1 and others 0) for the \( i \)-th sample. \( p_{ic} \) is the predicted probability that the \( i \)-th sample belongs to class \( c \). \( \alpha \) is a balancing factor that adjusts the weight between positive and negative samples. \( \gamma \) (gamma) is the focusing parameter, which controls the emphasis on hard-to-classify samples.  

\subsection{Limitation and Future Work}
The main limitation is our reliance on monocular RGB input. The model operates on per-frame inference, without leveraging temporal continuity, which may reduce robustness in cases of fast motion, occlusion, or scene changes.

Another limitation is that Butter is tailored for structured driving scenarios, and not optimized for general-purpose object detection. The evaluation code is available in our \textit{GitHub repository}, and results are summarized below. As shown in the table, it achieves a strong mAP@50 of 60.9 on MS COCO~\cite{lin2014microsoft}, surpassing many compact models but underperforming compared to top task-specific detectors like YOLOv12-S~\cite{Tian2025} and Hyper-YOLO-S~\cite{hyperyolo}.

Future work includes: (1) extending FAFCE to support spatiotemporal fusion, and (2) incorporating domain-adaptive modules for broader applicability. We will clearly reflect these in the final version. 

Recent advances in generative modeling~\cite{lu2023tf,lu2024mace,lu2024robust,li2025set,gao2024eraseanything}, multimodal fusion~\cite{shen2025amess}, and anomaly detection~\cite{zhang2025dconad,zhang2025dhmp,zhang2025frect} provide promising directions for strengthening representation learning in object detection. These methods excel at modeling structural detail, cross-modal correlation, and uncertainty—key factors for robust perception in autonomous driving. We plan to integrate such techniques into our architecture to improve generalization under limited supervision in the future. Furthermore, legal and ethical implications, especially for generative components, must be carefully addressed to ensure system-level safety~\cite{lin2025ssrn}.

\begin{table}[h!]
\centering
\rowcolors{2}{gray!10}{white}
\begin{tabular}{l l c c}
\toprule
\textbf{Method} & \textbf{Reference} & \textbf{mAP@50} & \textbf{\# Params (M)} \\
\midrule
Gold-YOLO-N~\cite{wang2023gold}     & NIPS 23     & 55.7  & 5.6 \\
YOLOv8-N~\cite{glenn2023yolov8}        & GitHub 23   & 52.6  & 3.2 \\
YOLOv9-T~\cite{wang2024yolov9}        & ECCV 2024   & 53.1  & \textbf{2.0} \\
Hyper-YOLO-T~\cite{hyperyolo}    & TPAMI 24    & 54.5  & 3.1 \\
Hyper-YOLO-N~\cite{hyperyolo}    & TPAMI 24    & 58.3  & 4.0 \\
Hyper-YOLO-S~\cite{hyperyolo}    & TPAMI 24    & \textbf{65.1}  & 14.8 \\
YOLOv12-N~\cite{Tian2025}       & arXiv 25    & 56.7  & 2.6 \\
YOLOv12-S~\cite{Tian2025}       & arXiv 25    & 65.0  & 9.3 \\
\textbf{Butter} & -           & 60.9  & 5.4 \\
\bottomrule
\end{tabular}
\caption{Table: The performance comparison between Butter and other methods on MS COCO~\cite{lin2014microsoft}.}
\end{table}

\section{Additional Results}
\subsection{Visualization of FAFCE Attention Enhancement}

\begin{figure*}[h]
    \centering
    \captionsetup[subfloat]{labelsep=none,format=plain,labelformat=empty,textfont={normalfont},font={normalfont,Huge}}
    \subfloat{\includegraphics[width=0.4\textwidth]{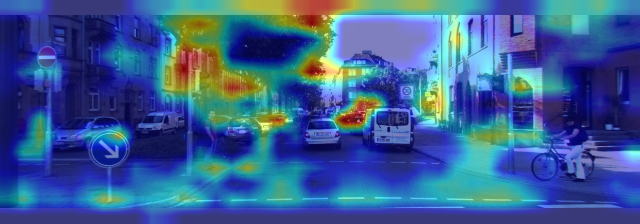}}  \hspace{0.2mm}
    \subfloat{\includegraphics[width=0.4\textwidth]{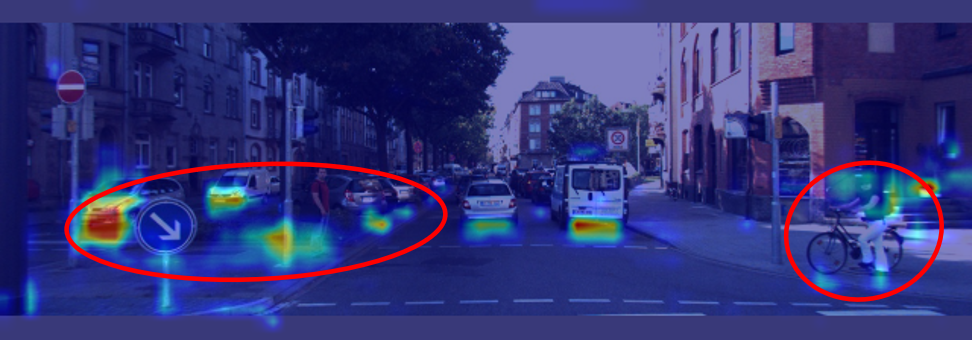}}  \\
    \vspace{-3mm}
    \subfloat{\includegraphics[width=0.4\textwidth]{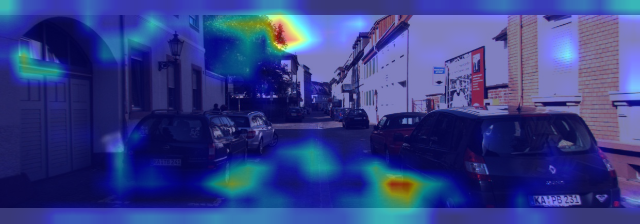}}  \hspace{0.2mm}
    \subfloat{\includegraphics[width=0.4\textwidth]{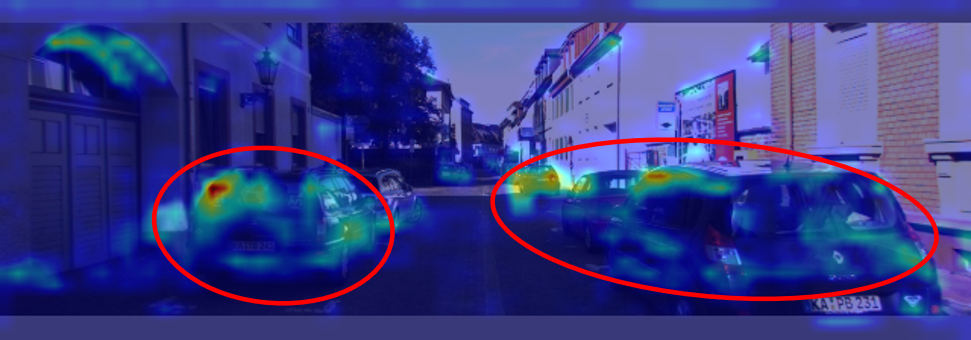}}  \\
    \vspace{-3mm}
    \subfloat{\includegraphics[width=0.4\textwidth]{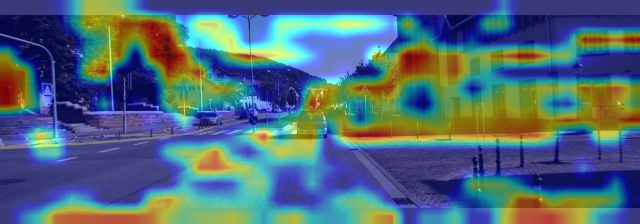}}  \hspace{0.2mm}
    \subfloat{\includegraphics[width=0.4\textwidth]{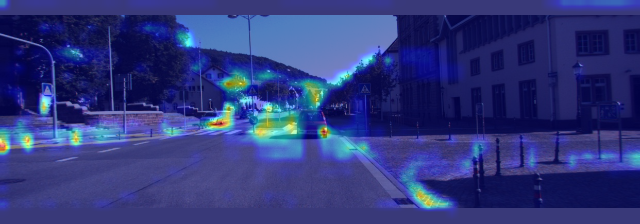}}  \\
    \vspace{-3mm}
    \subfloat{\includegraphics[width=0.4\textwidth]{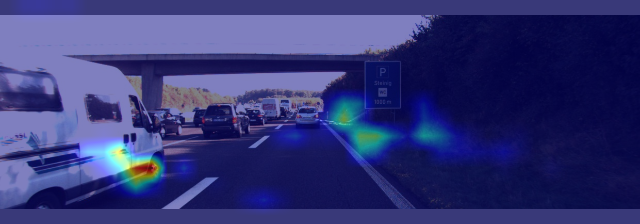}}  \hspace{0.2mm}
    \subfloat{\includegraphics[width=0.4\textwidth]{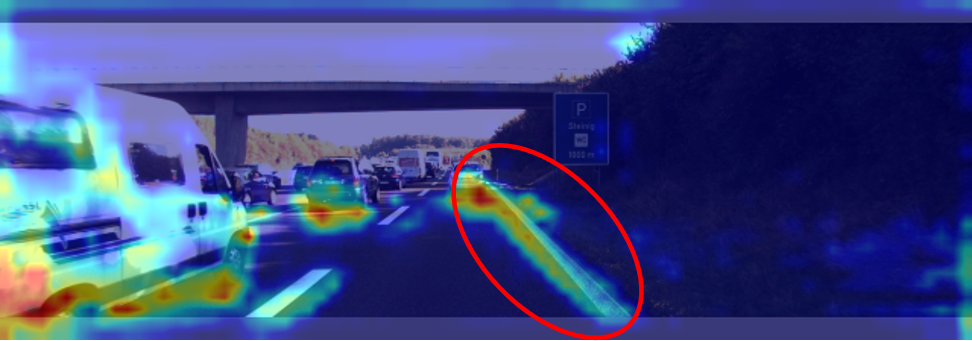}}  \\
    \vspace{-3mm}
    \subfloat{\includegraphics[width=0.4\textwidth]{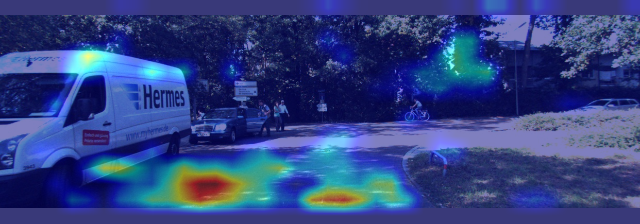}}  \hspace{0.2mm}
    \subfloat{\includegraphics[width=0.4\textwidth]{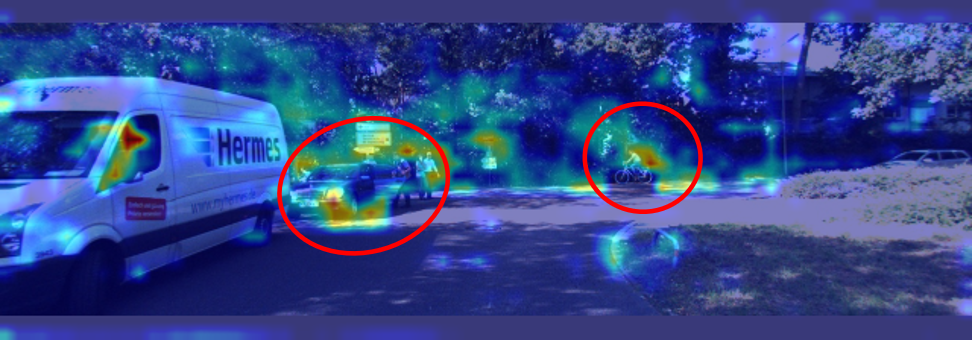}}  \\
    \vspace{-3mm}
    \subfloat{\includegraphics[width=0.4\textwidth]{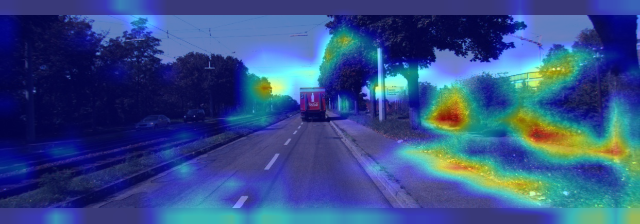}}  \hspace{0.2mm}
    \subfloat{\includegraphics[width=0.4\textwidth]{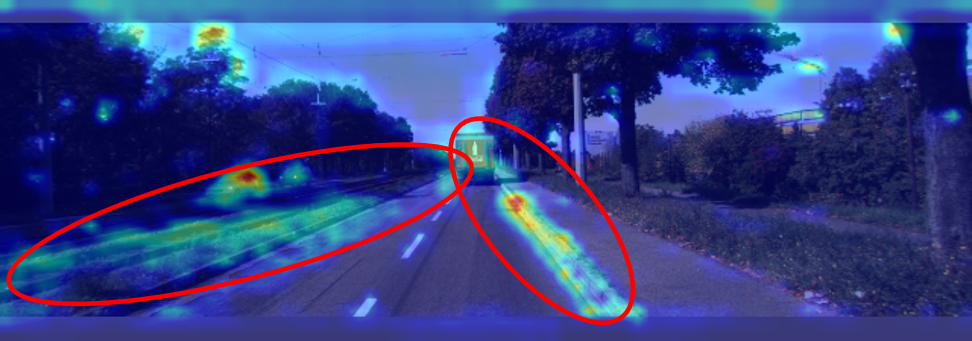}}  \\
    \vspace{-3mm}
    \subfloat{\includegraphics[width=0.4\textwidth]{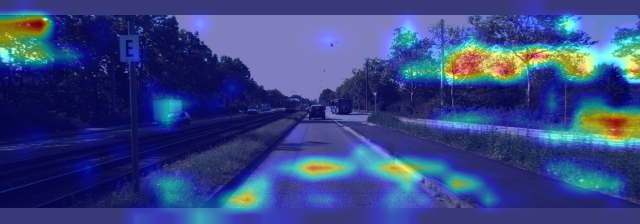}}  \hspace{0.2mm}
    \subfloat{\includegraphics[width=0.4\textwidth]{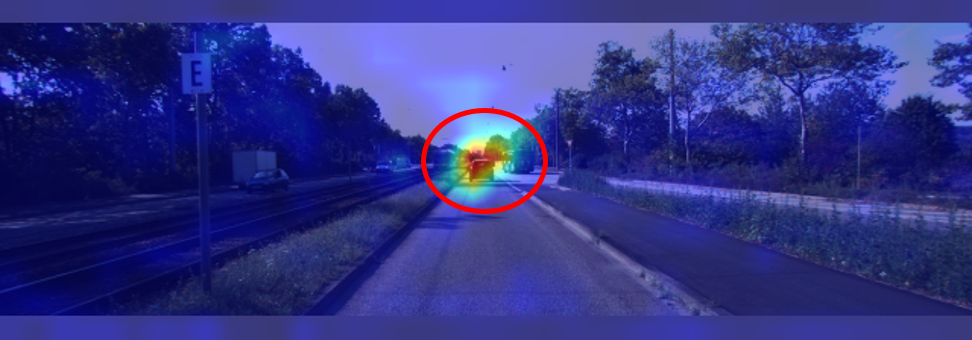}}  \\
    \vspace{-3mm}
    \subfloat[without FAFCE]{\includegraphics[width=0.4\textwidth]{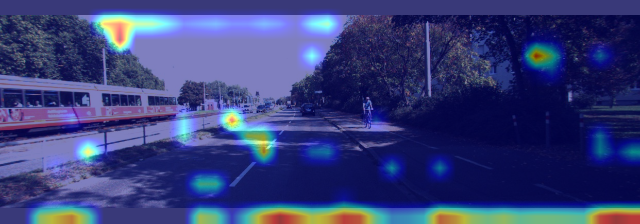}}  \hspace{0.2mm}
    \subfloat[with FAFCE]{\includegraphics[width=0.4\textwidth]{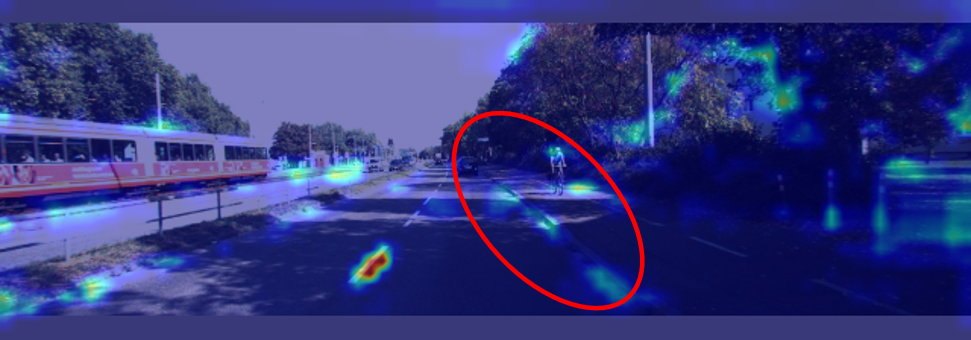}}  \\
    \caption{Heatmap Comparison of Model Attention in Butter.}
    \label{fig:additonal-htmp-1}

\end{figure*}

\begin{figure*}[h]
    \centering
    \captionsetup[subfloat]{labelsep=none,format=plain,labelformat=empty,font=Huge,textfont={normalfont}}
    \subfloat{\includegraphics[width=0.4\textwidth]{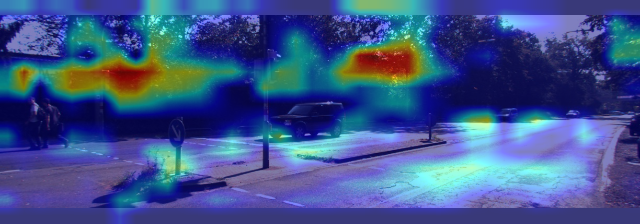}}  \hspace{0.2mm}
    \subfloat{\includegraphics[width=0.4\textwidth]{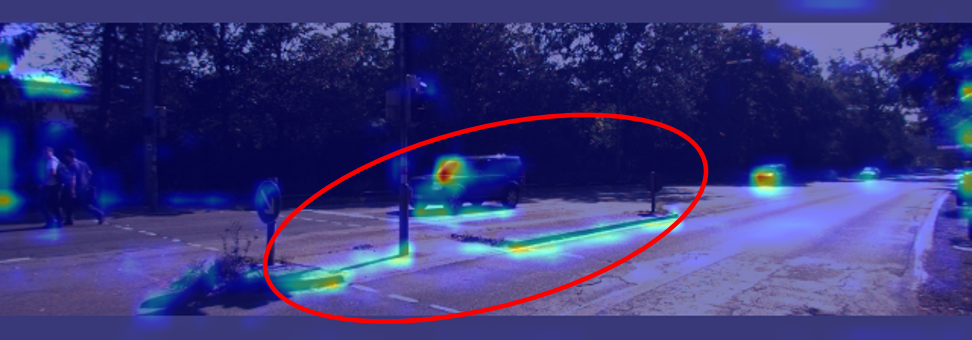}}  \\
    \vspace{-3mm}
    \subfloat{\includegraphics[width=0.4\textwidth]{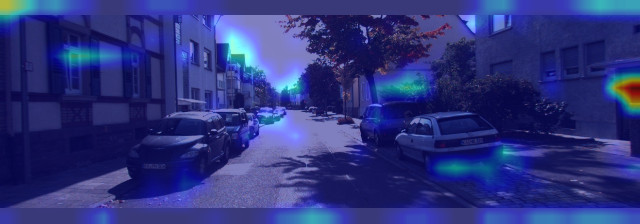}}  \hspace{0.2mm}
    \subfloat{\includegraphics[width=0.4\textwidth]{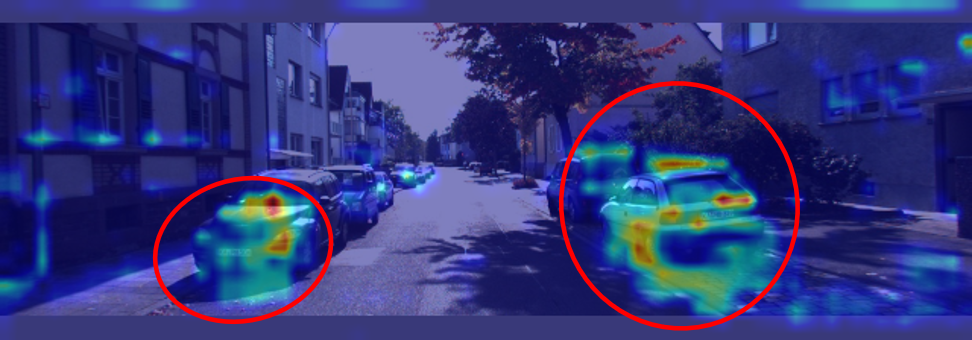}}  \\
    \vspace{-3mm}
    \subfloat{\includegraphics[width=0.4\textwidth]{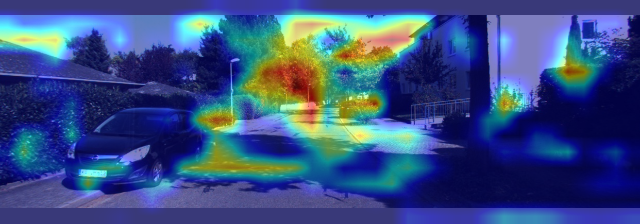}}  \hspace{0.2mm}
    \subfloat{\includegraphics[width=0.4\textwidth]{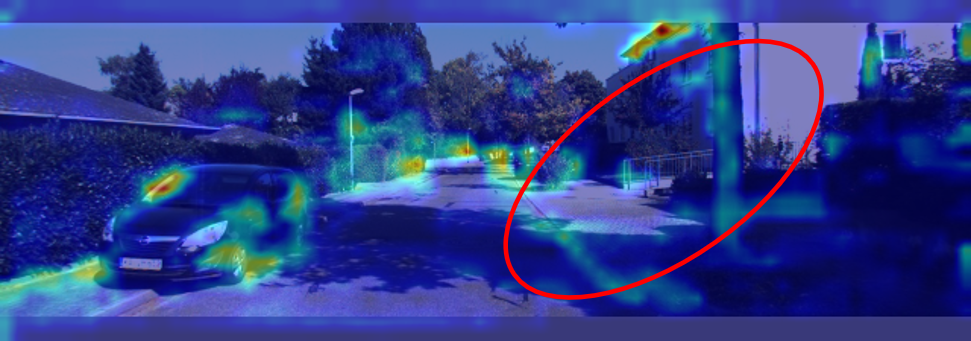}}  \\
    \vspace{-3mm}
    \subfloat{\includegraphics[width=0.4\textwidth]{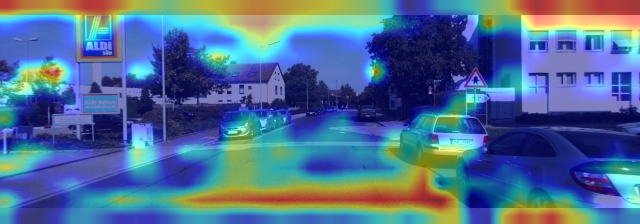}}  \hspace{0.2mm}
    \subfloat{\includegraphics[width=0.4\textwidth]{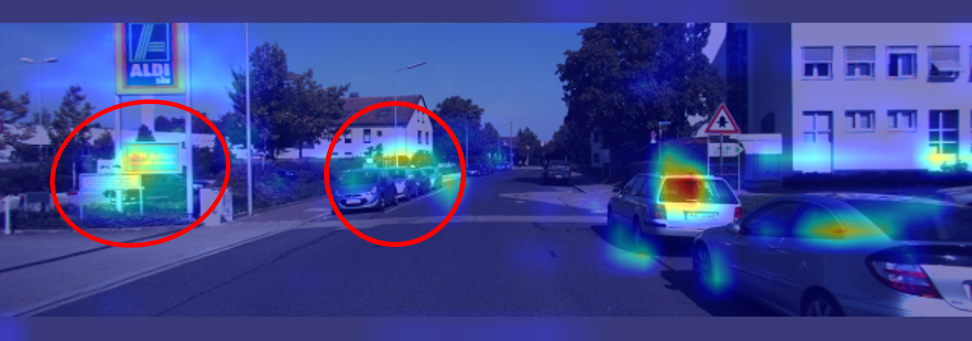}}  \\
    \vspace{-3mm}
    \subfloat{\includegraphics[width=0.4\textwidth]{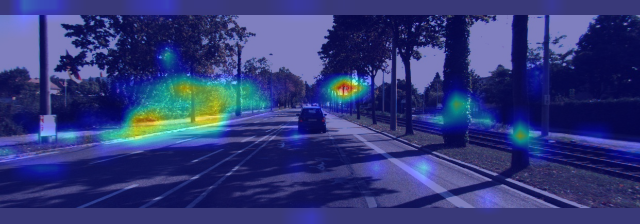}}  \hspace{0.2mm}
    \subfloat{\includegraphics[width=0.4\textwidth]{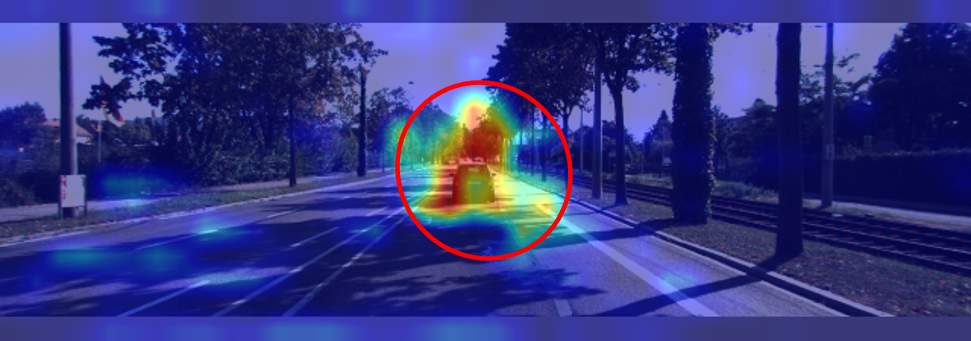}}  \\
    \vspace{-3mm}
    \subfloat{\includegraphics[width=0.4\textwidth]{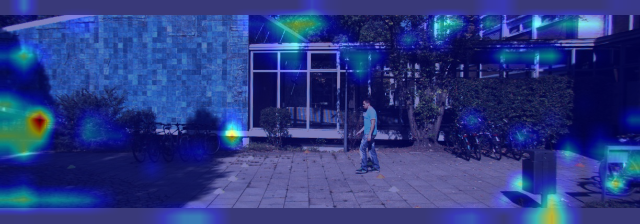}}  \hspace{0.2mm}
    \subfloat{\includegraphics[width=0.4\textwidth]{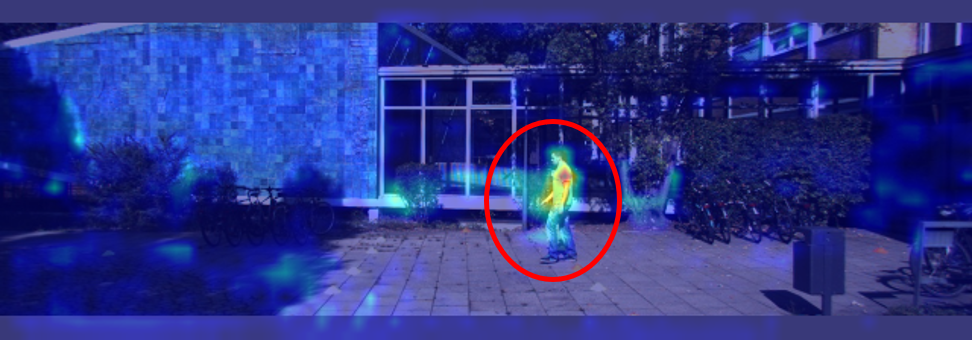}}  \\
    \vspace{-3mm}
    \subfloat{\includegraphics[width=0.4\textwidth]{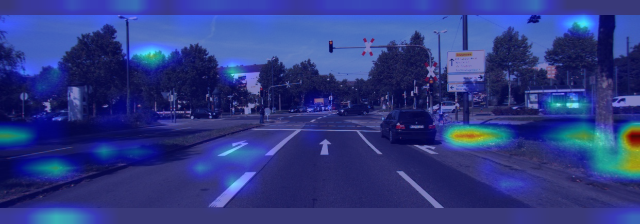}}  \hspace{0.2mm}
    \subfloat{\includegraphics[width=0.4\textwidth]{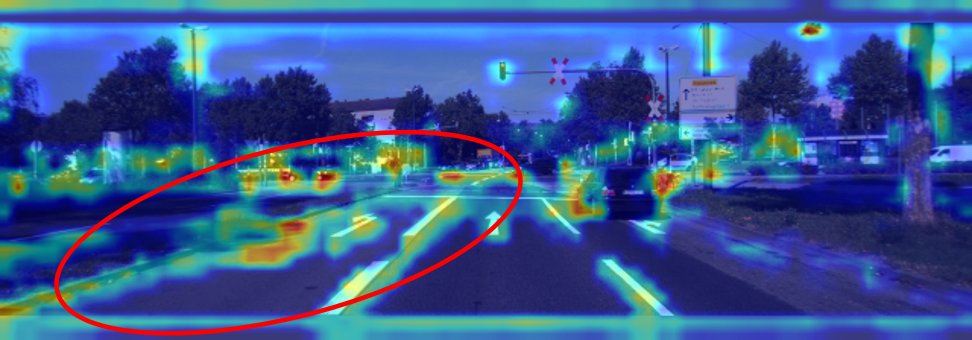}}  \\
    \vspace{-3mm}
    \subfloat[without FAFCE]{\includegraphics[width=0.4\textwidth]{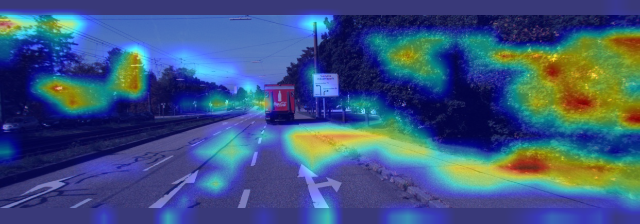}}  \hspace{0.2mm}
    \subfloat[with FAFCE]{\includegraphics[width=0.4\textwidth]{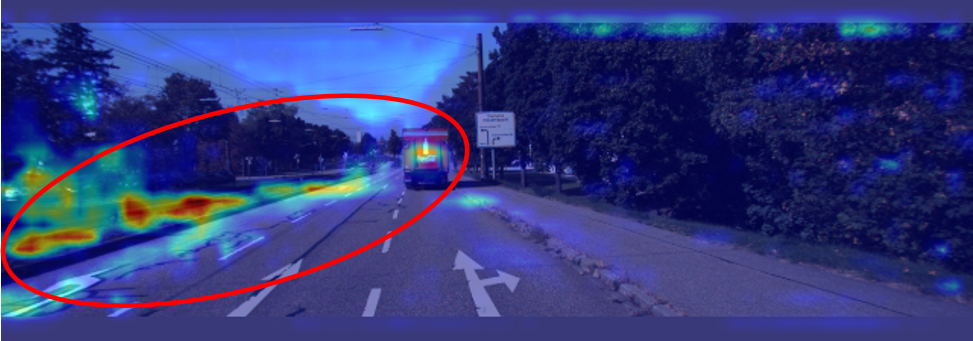}}  \\
    \caption{Heatmap Comparison of Model Attention in Butter.}
    \label{fig:additonal-htmp-2}

\end{figure*}

To better illustrate the effect of FAFCE, we visualize additional heatmaps of the model's attention. All cases are selected from the KITTI~\cite{geiger2012ready} dataset. As shown in Fig.~\ref{fig:additonal-htmp-1} and Fig.~\ref{fig:additonal-htmp-2}, after the application of FAFCE, attention is more effectively directed towards targets with specific semantic content, such as lane markings, vehicles, pedestrians, and traffic signs. This observation suggests that FAFCE integrates low-level details with high-level semantic information, thereby significantly enhancing the model's capacity for hierarchical representation learning.

\subsection{Visualization of Object Detection Task of Butter}

\begin{figure*}[h]
    \centering
    \subfloat{\includegraphics[width=0.3\textwidth]{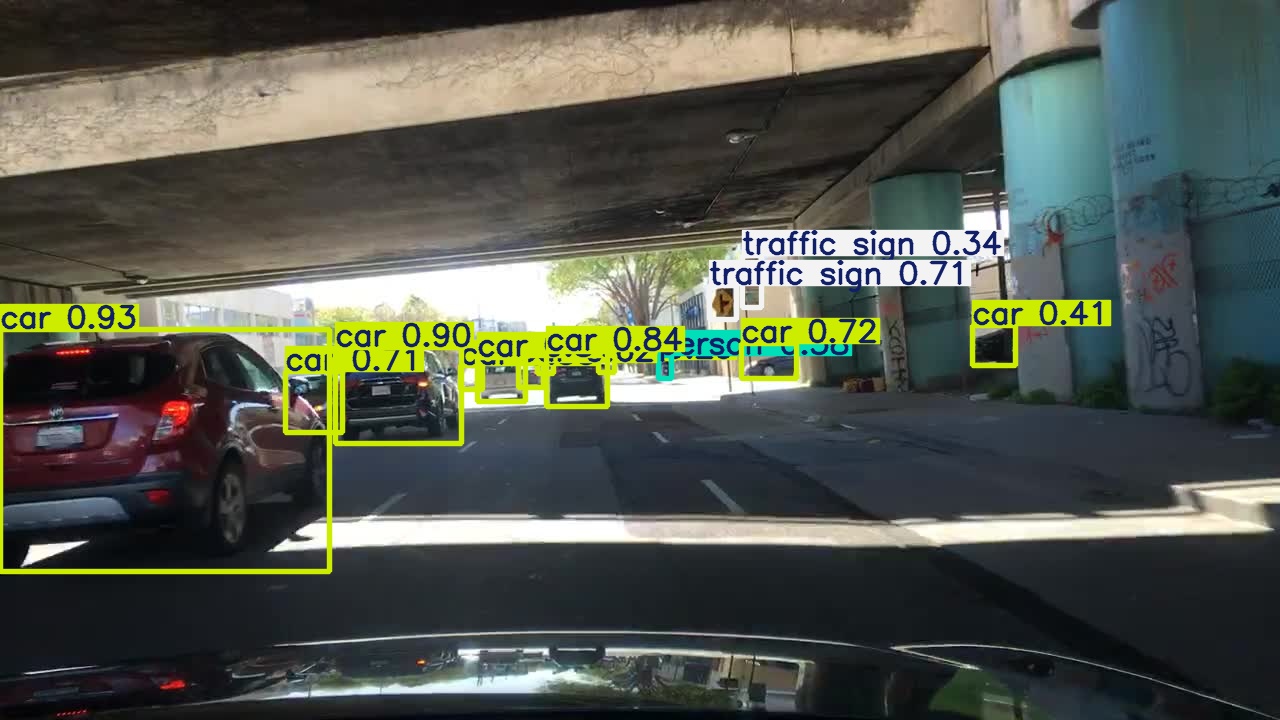}}  \hspace{0.2mm}
    \subfloat{\includegraphics[width=0.3\textwidth]{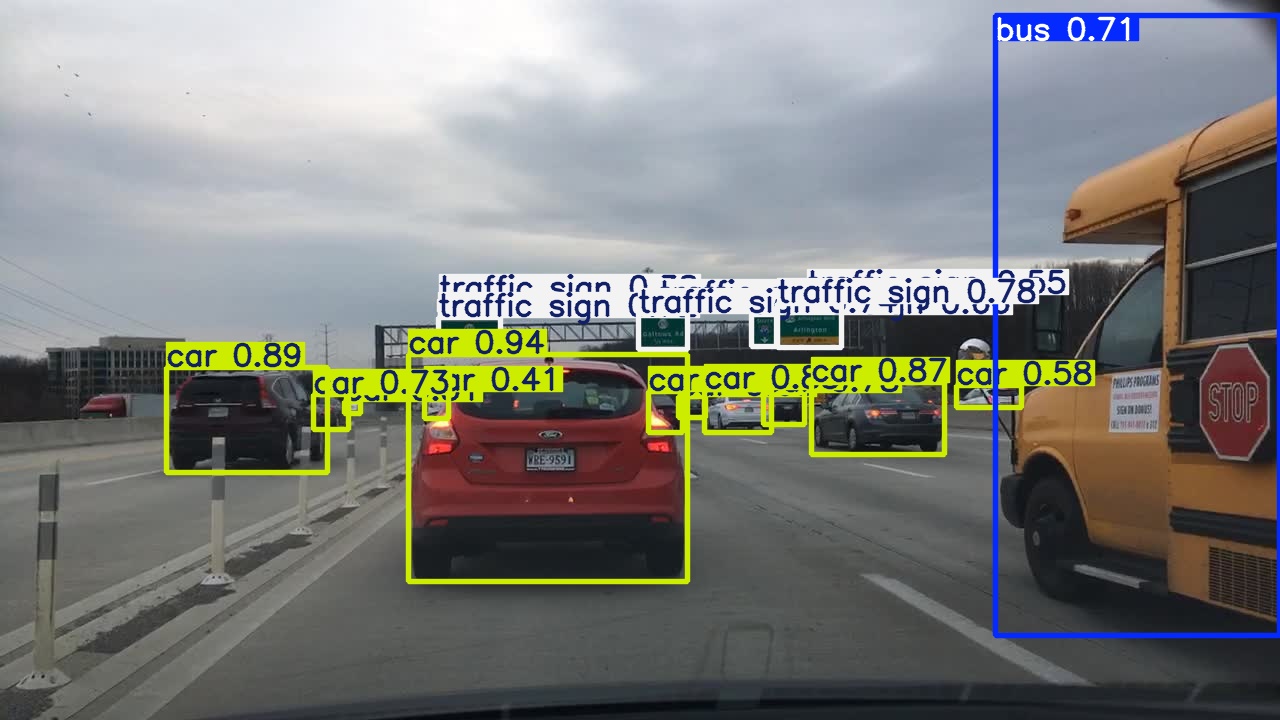}}  \hspace{0.2mm}
    \subfloat{\includegraphics[width=0.3\textwidth]{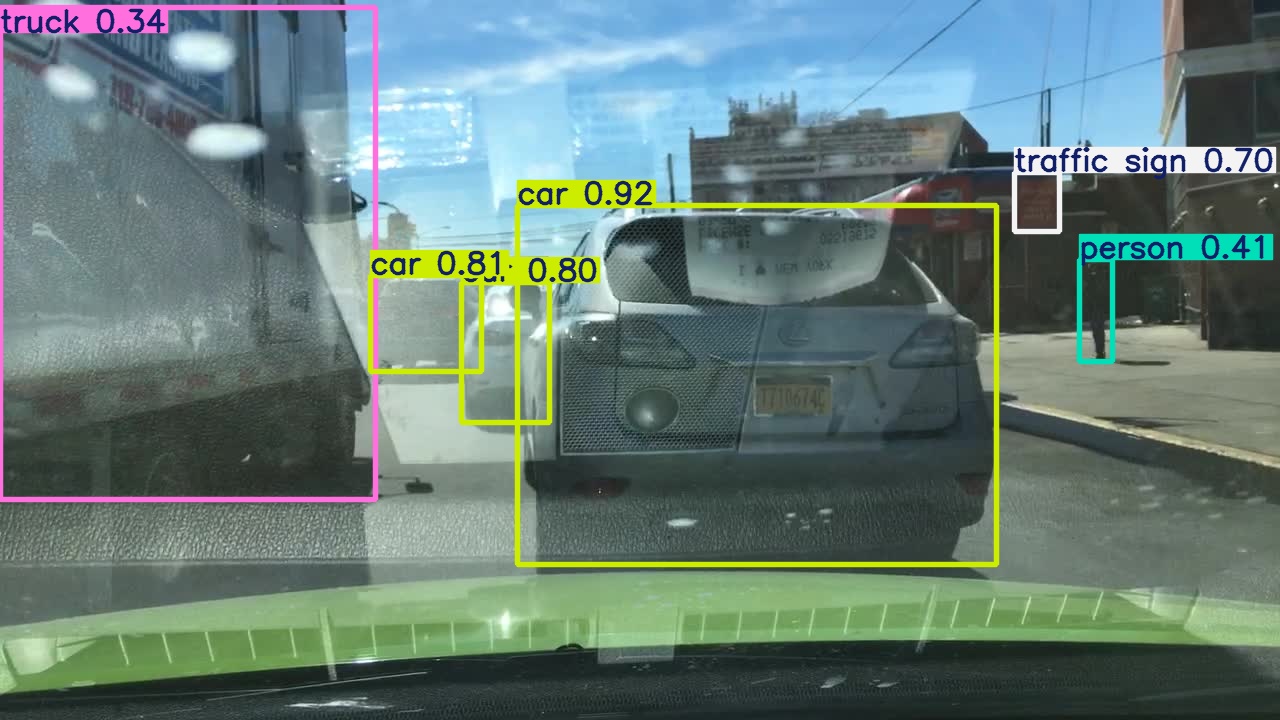}} \\
    \vspace{-3mm}
    \subfloat{\includegraphics[width=0.3\textwidth]{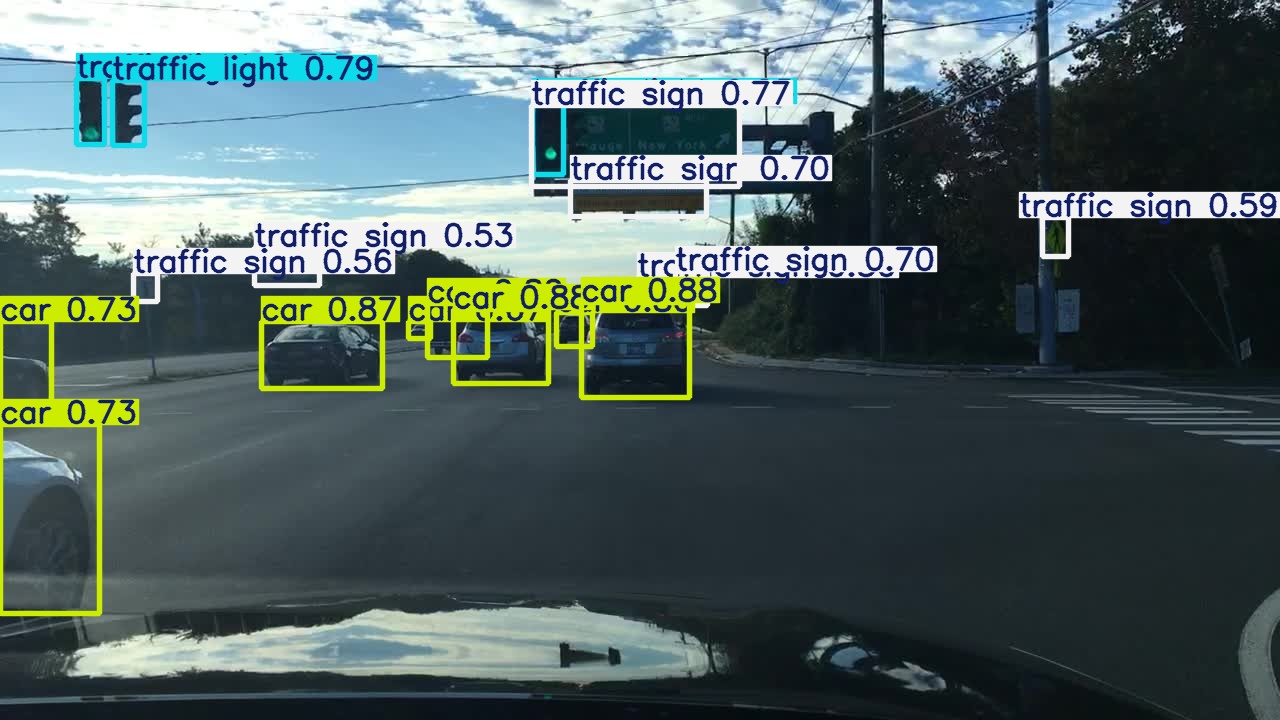}}  \hspace{0.2mm}
    \subfloat{\includegraphics[width=0.3\textwidth]{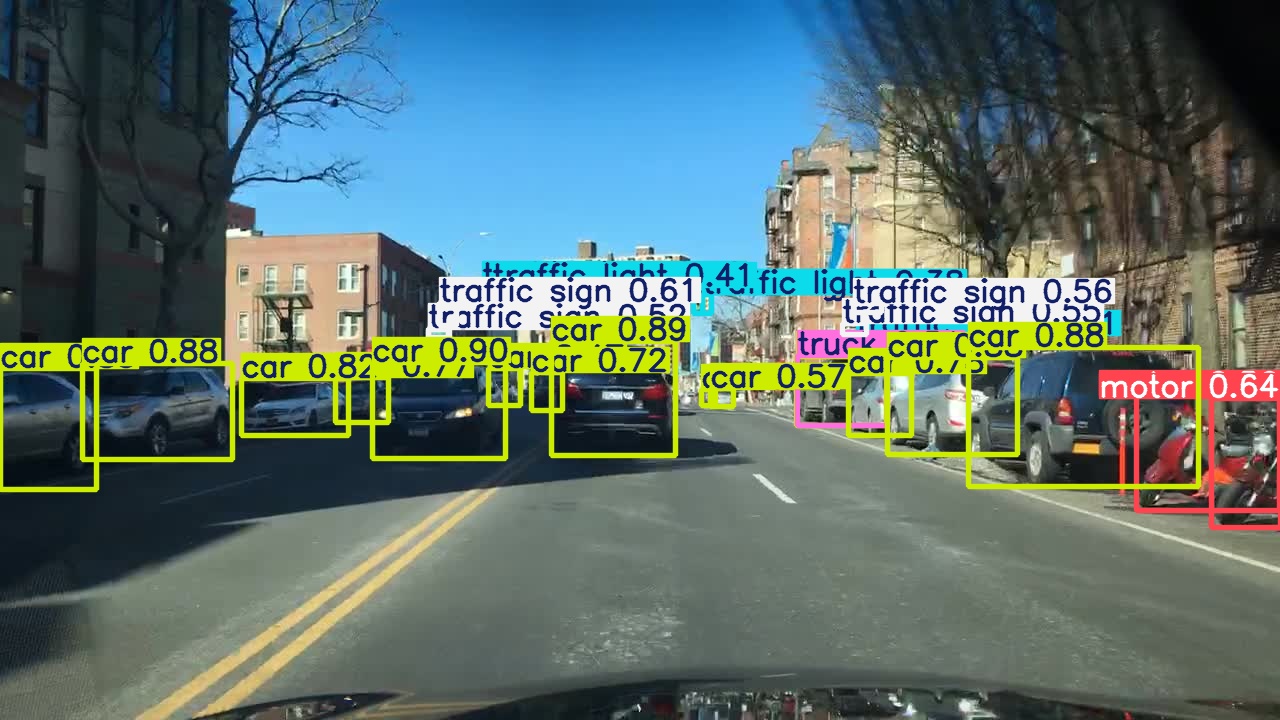}}  \hspace{0.2mm}
    \subfloat{\includegraphics[width=0.3\textwidth]{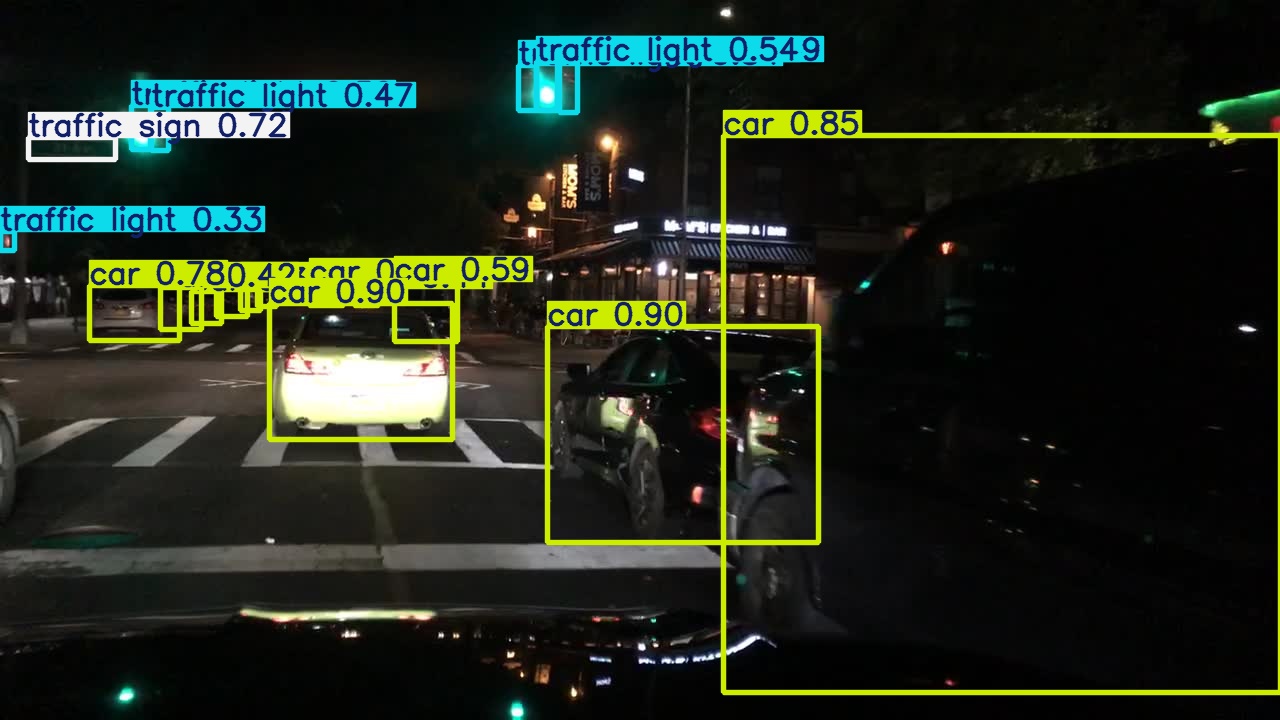}}
    \caption{Visualization of Object Detection Task of Butter in BDD100K Dataset.}
    \label{fig:case-bdd100k}

\end{figure*}

\begin{figure*}[h]
    \centering
    \subfloat{\includegraphics[width=0.3\textwidth]{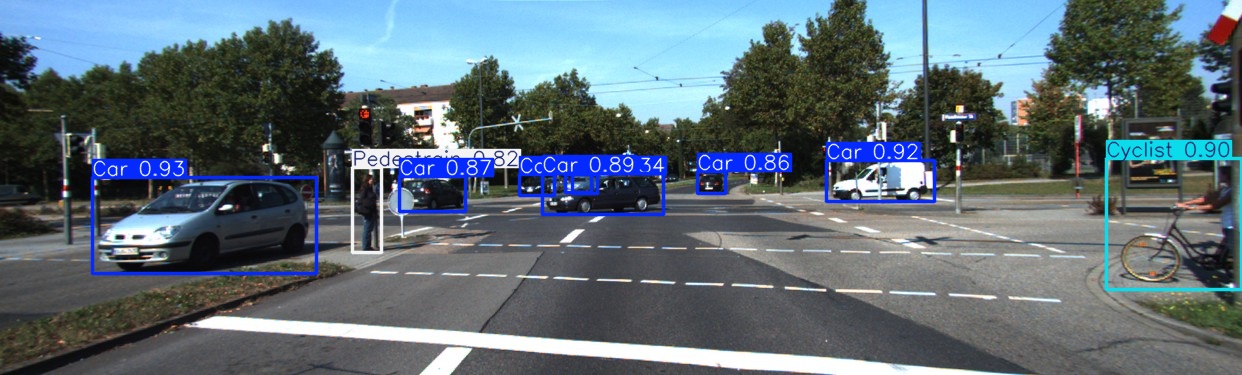}} \hspace{0.2mm}
    \subfloat{\includegraphics[width=0.3\textwidth]{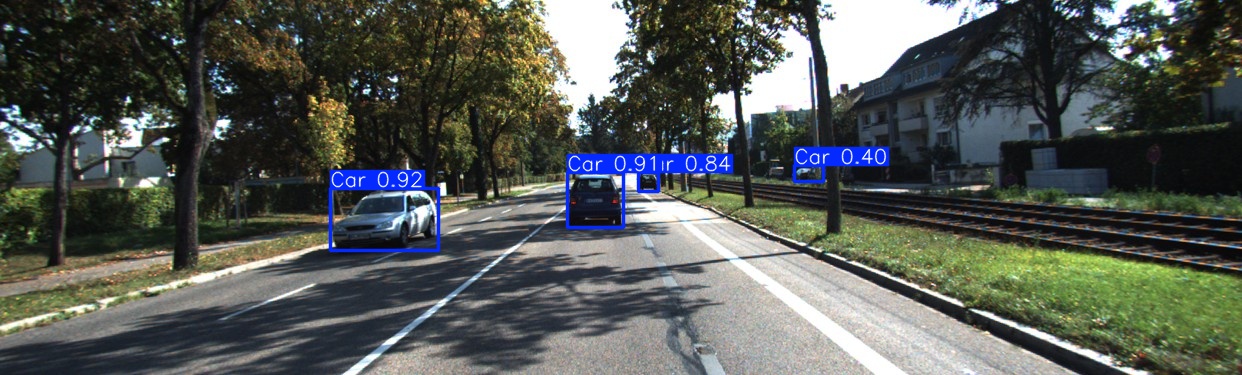}}  \hspace{0.2mm}
    \subfloat{\includegraphics[width=0.3\textwidth]{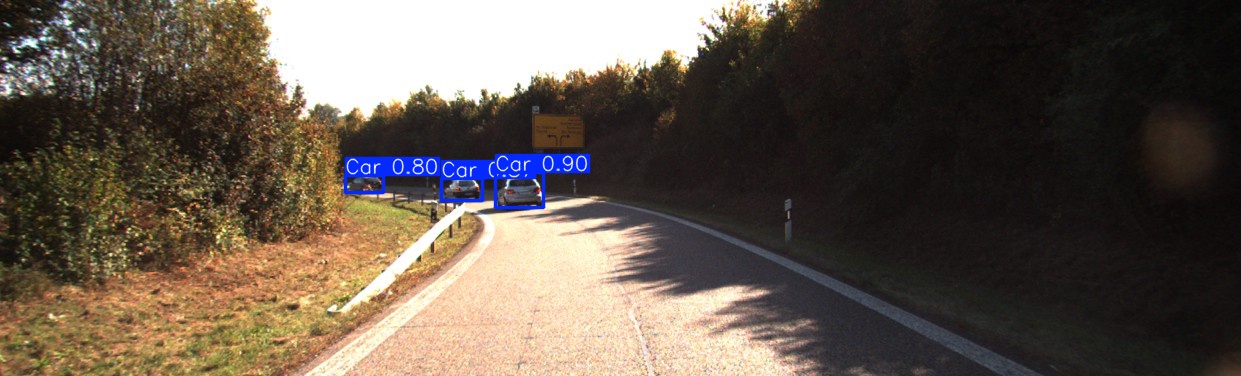}}\\
    \vspace{-3mm}
    \subfloat{\includegraphics[width=0.3\textwidth]{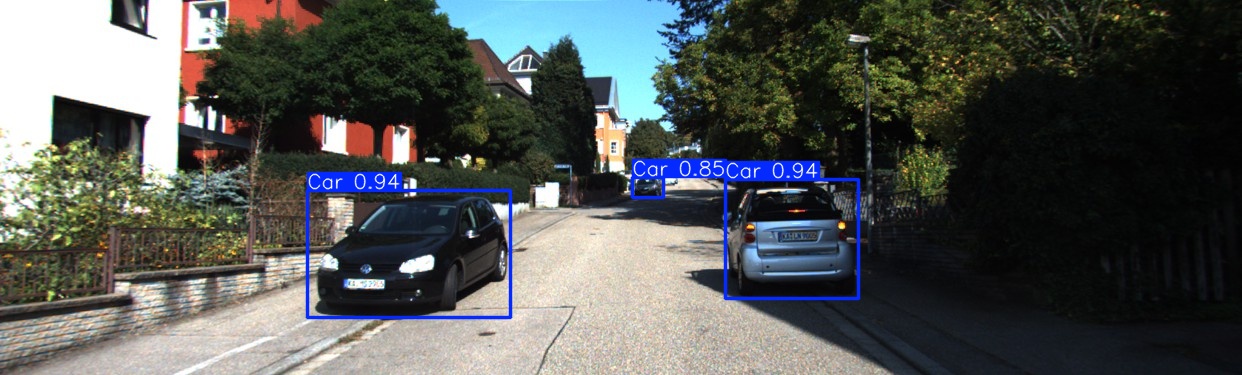}}  \hspace{0.2mm}
    \subfloat{\includegraphics[width=0.3\textwidth]{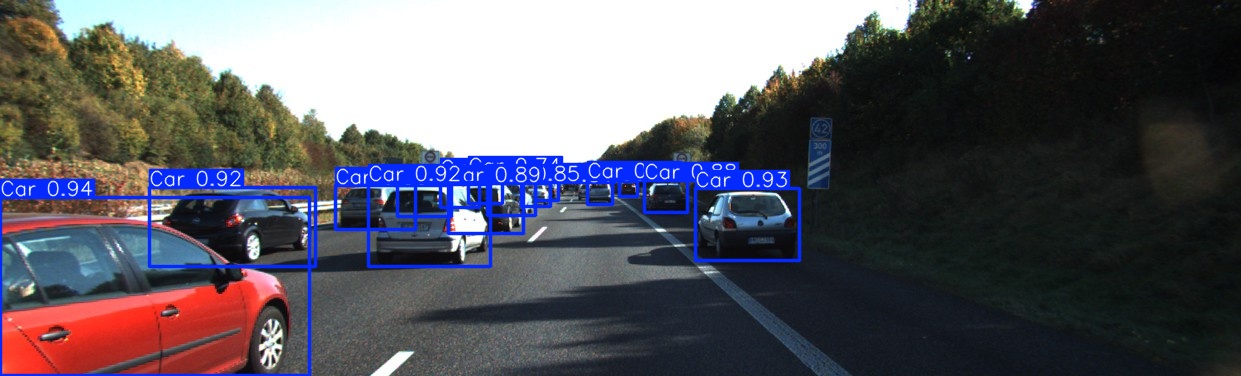}}  \hspace{0.2mm}
    \subfloat{\includegraphics[width=0.3\textwidth]{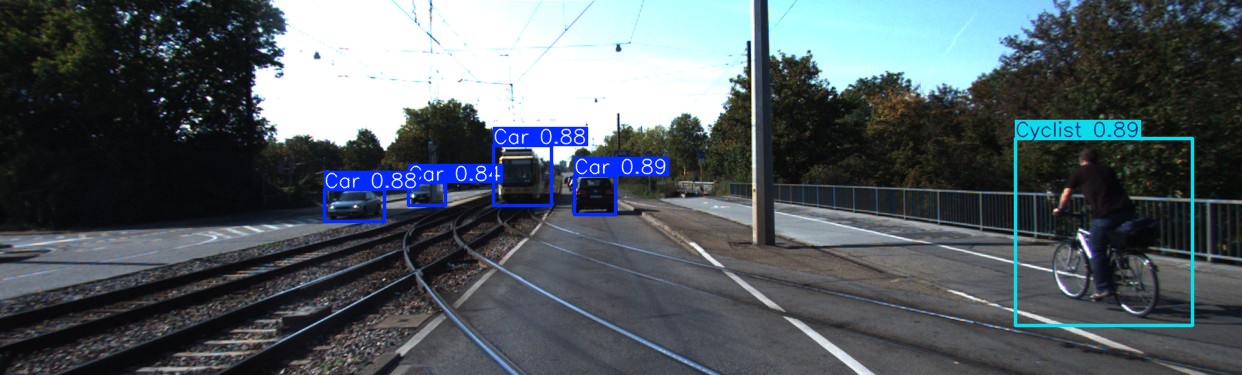}}
    \caption{Visualization of Object Detection Task of Butter in KITTI Dataset.}
\label{fig:case-kitti}
\end{figure*}

\begin{figure*}[h]
    \centering
    \subfloat{\includegraphics[width=0.3\textwidth]{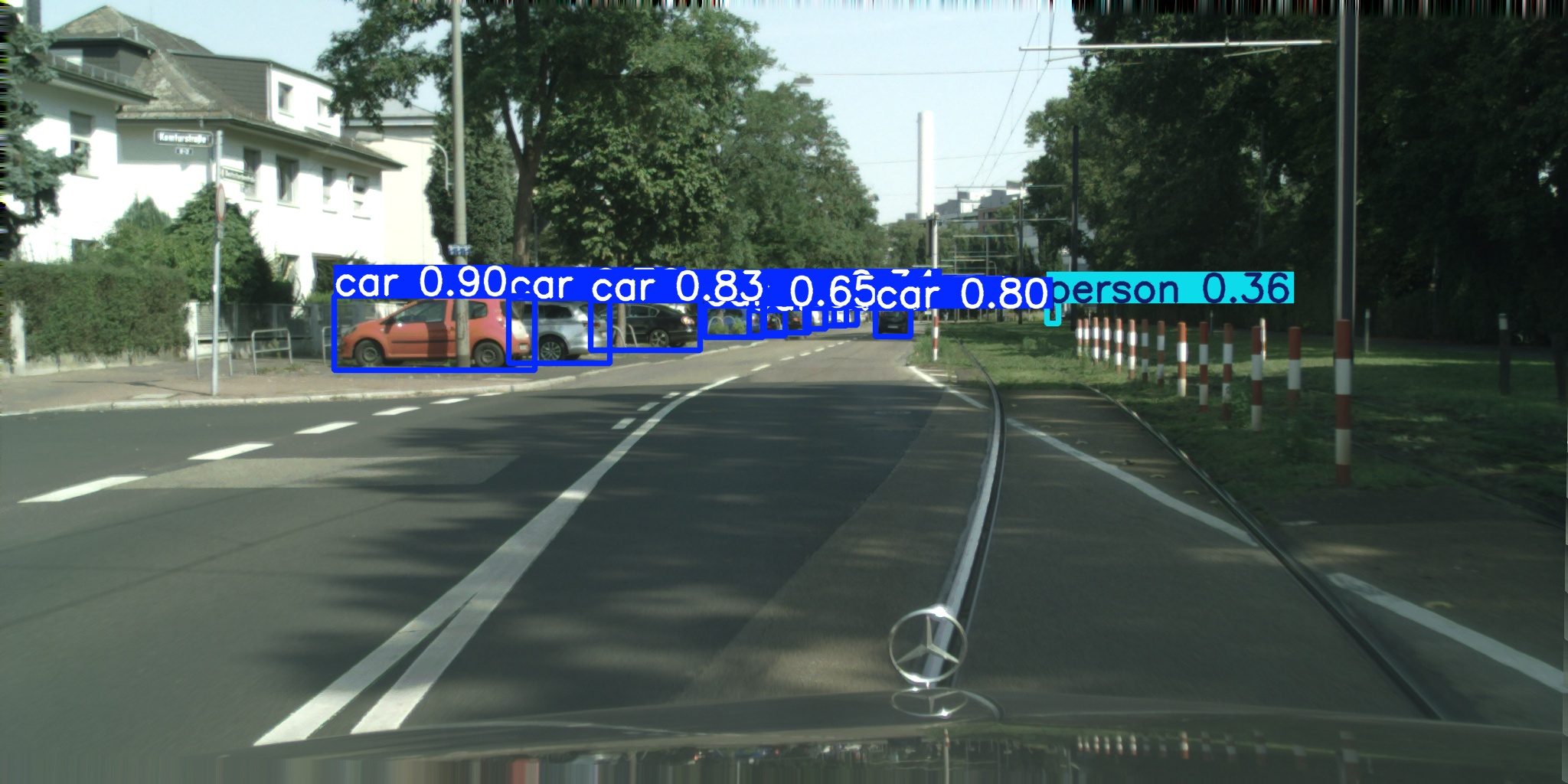}}  \hspace{0.2mm}
    \subfloat{\includegraphics[width=0.3\textwidth]{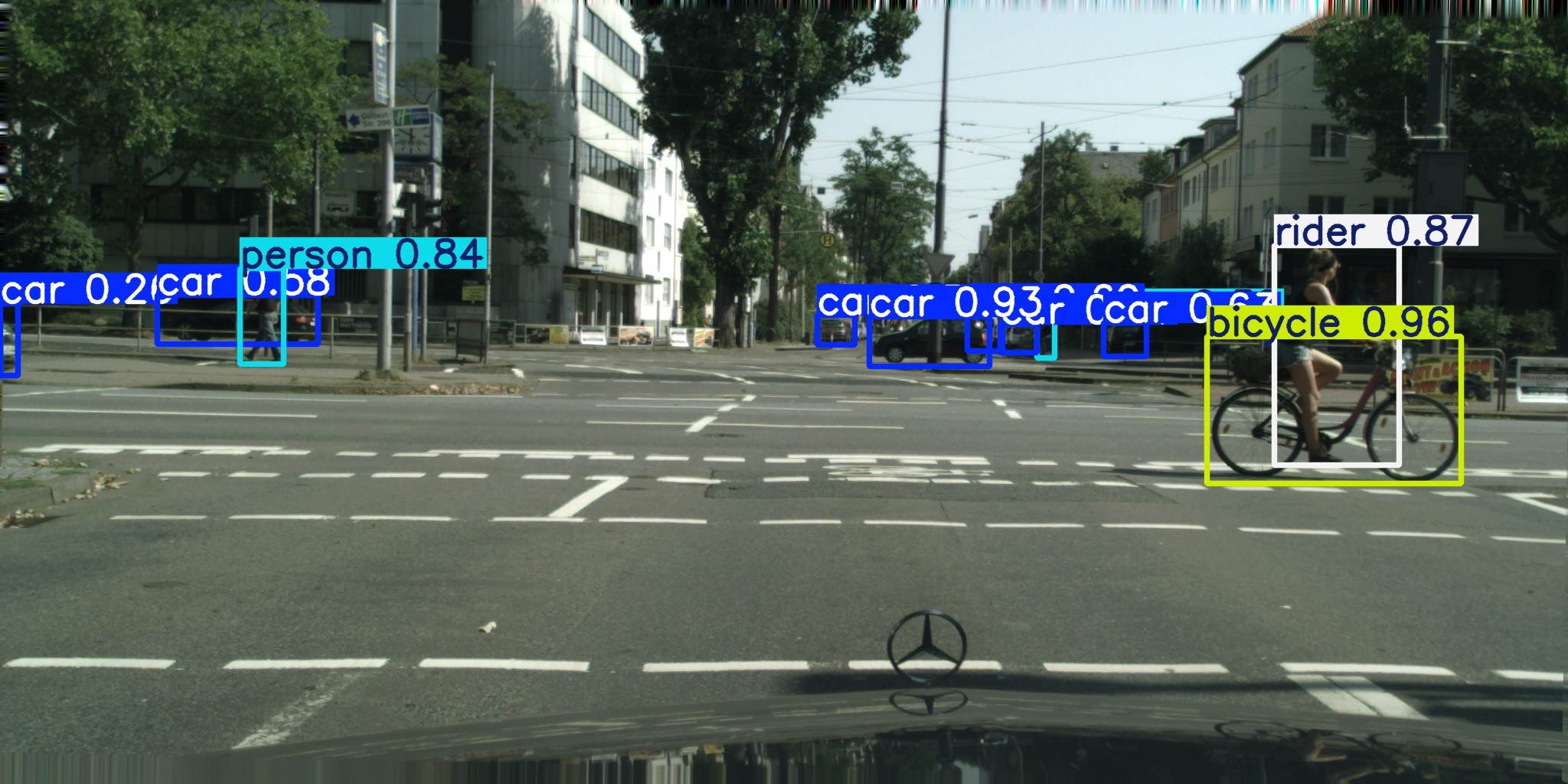}}  \hspace{0.2mm}
    \subfloat{\includegraphics[width=0.3\textwidth]{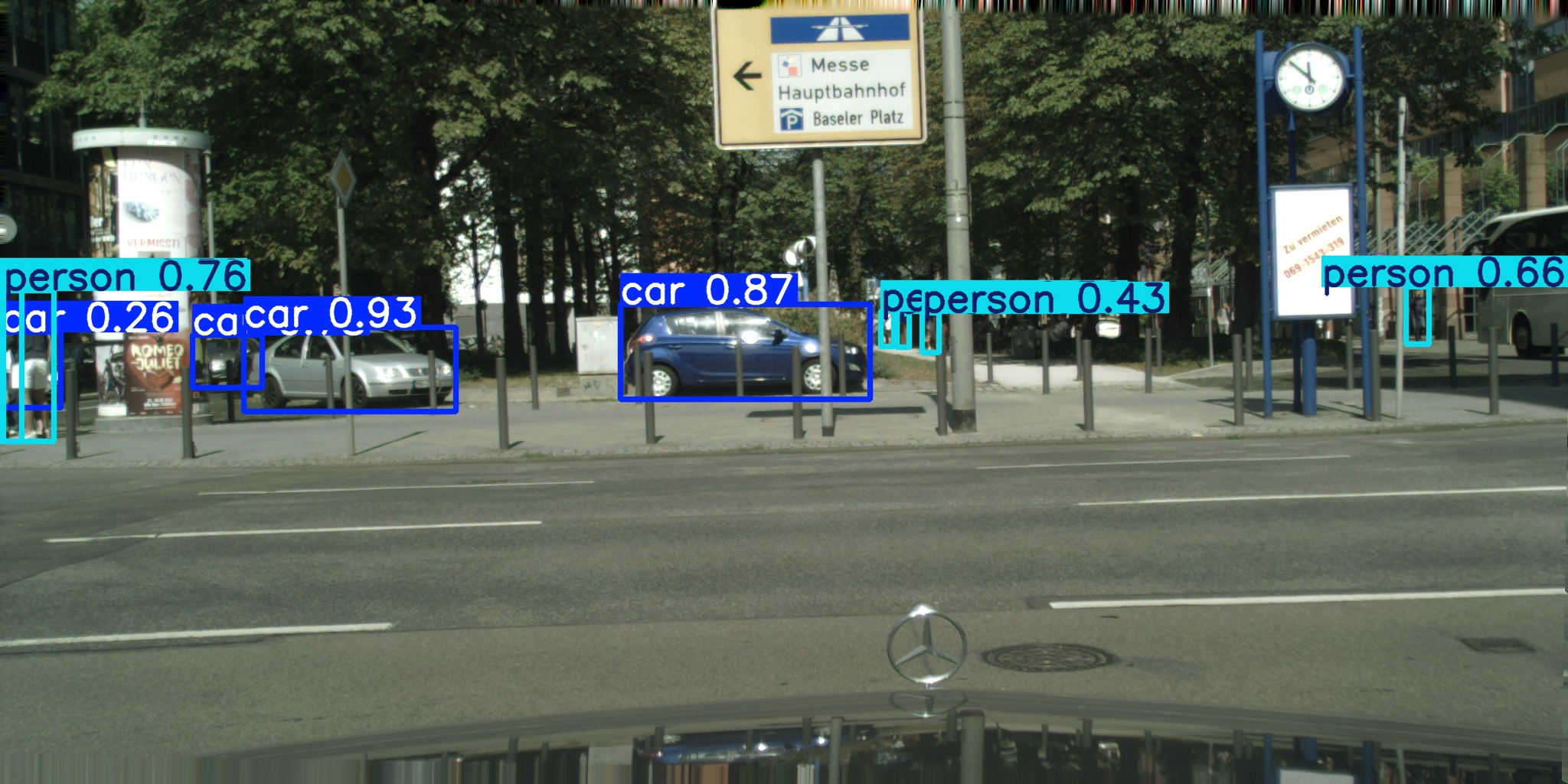}} \\
    \vspace{-3mm}
    \subfloat{\includegraphics[width=0.3\textwidth]{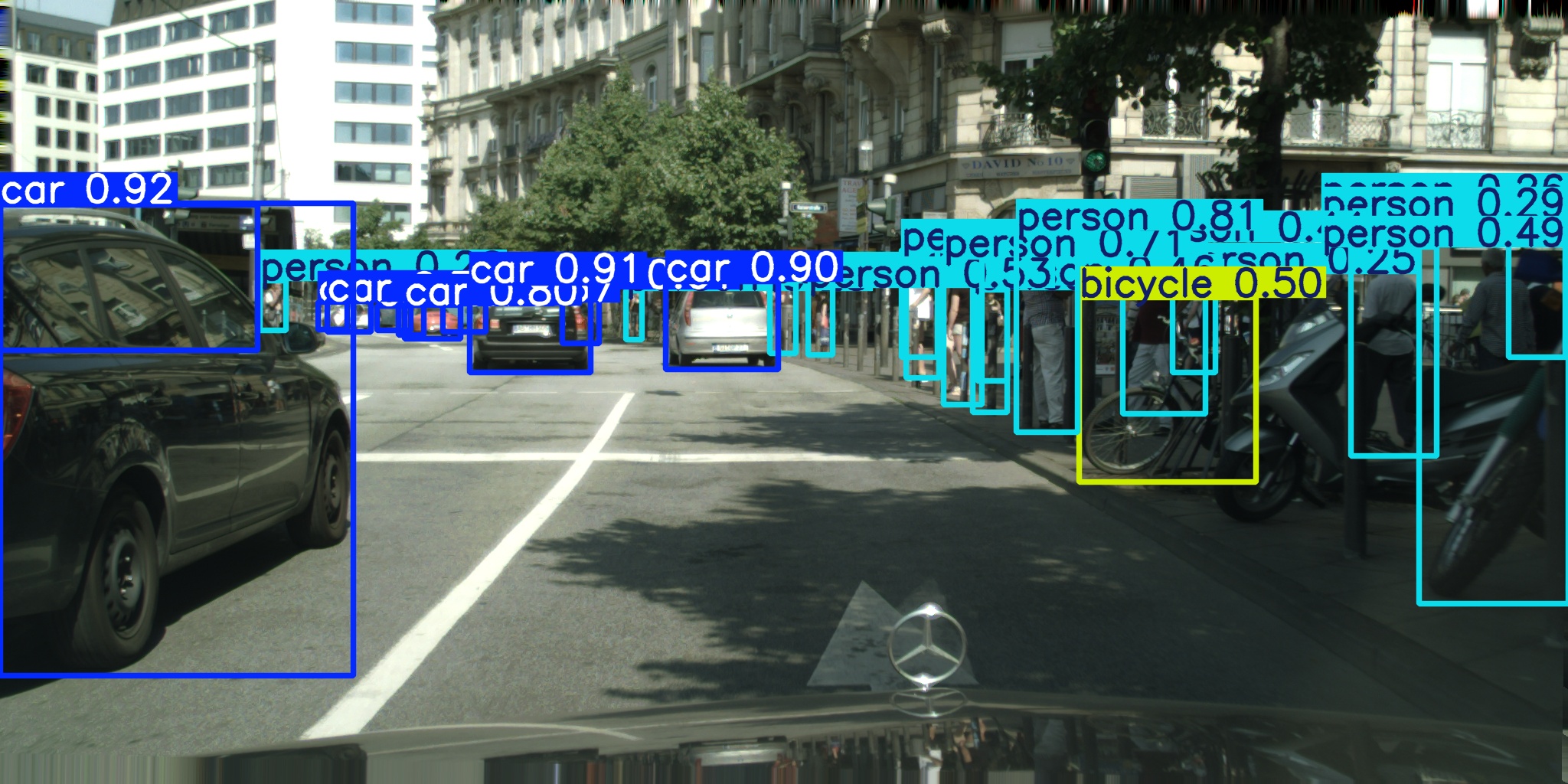}}  \hspace{0.2mm}
    \subfloat{\includegraphics[width=0.3\textwidth]{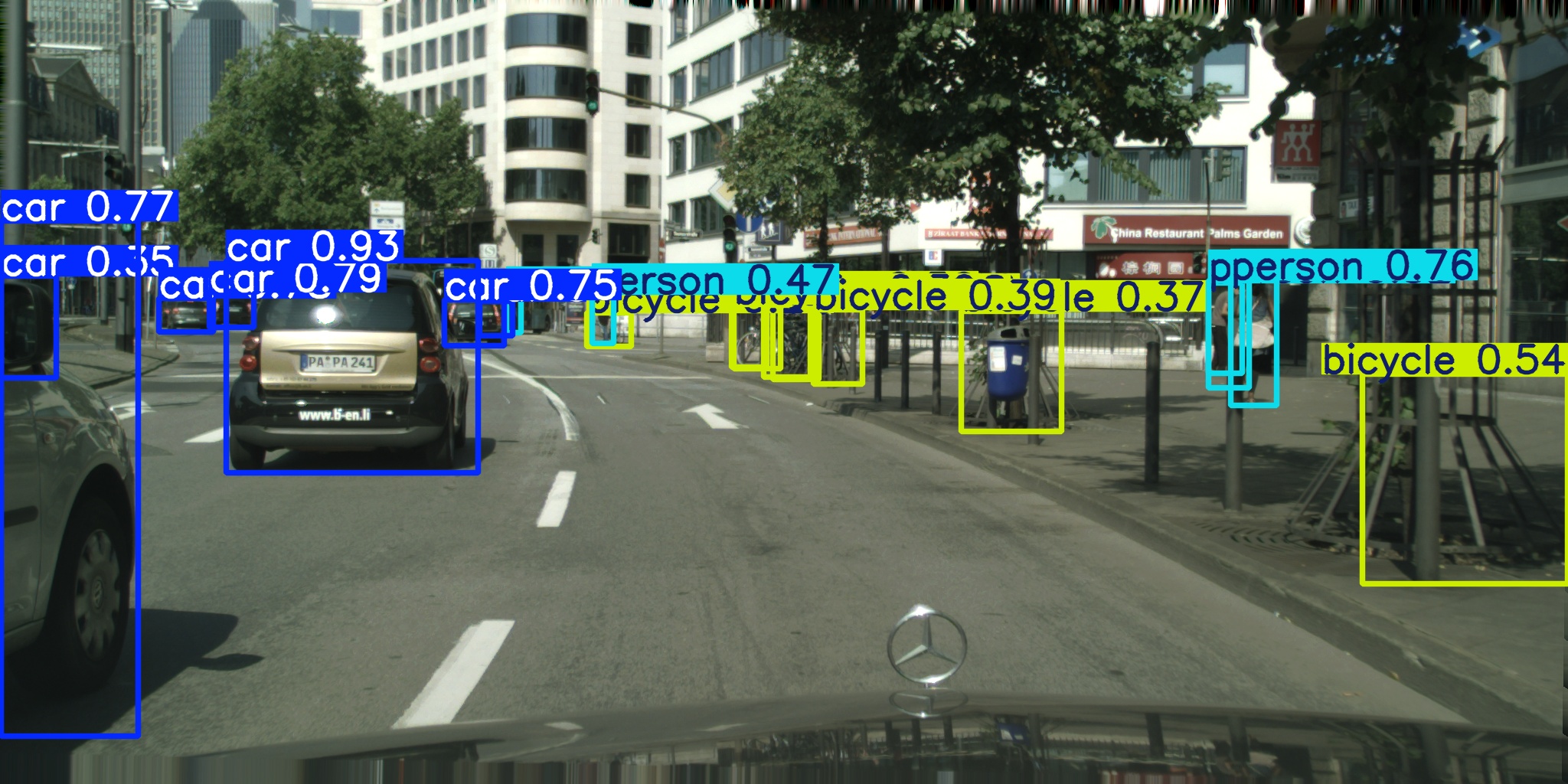}}  \hspace{0.2mm}
    \subfloat{\includegraphics[width=0.3\textwidth]{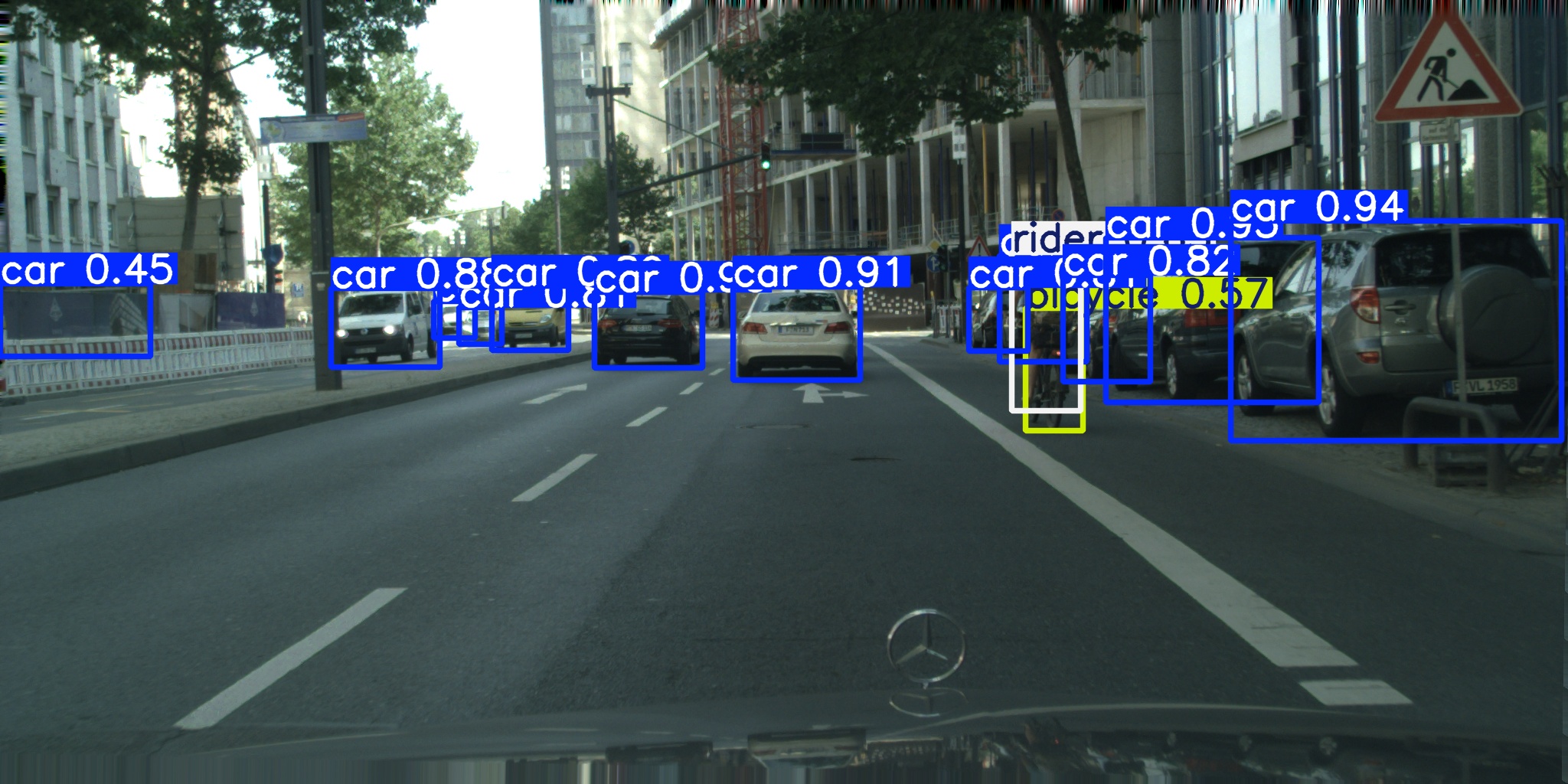}}
    \caption{Visualization of Object Detection Task of Butter in Cityscapes Dataset.}
    \label{fig:case-cityscapes}
\end{figure*}

We select several samples from the KITTI (Fig.~\ref{fig:case-kitti}), BDD100K (Fig.~\ref{fig:case-bdd100k}), and Cityscapes (Fig.~\ref{fig:case-cityscapes}) datasets to showcase the detection performance of Butter. Butter demonstrates its ability to detect a wide range of objects, even those that are small or densely overlapped. These predictions further substantiate the outstanding detection performance in the autonomous driving scenario of Butter.



\end{document}